\documentclass{article}
\usepackage{colm_files/colm2024_conference}
\usepackage{amsmath}
\usepackage{microtype}
\usepackage[pagebackref,breaklinks,colorlinks,citecolor=blue]{hyperref}
\usepackage{subcaption}
\usepackage{url}
\usepackage{booktabs}
\usepackage{todonotes}
\usepackage{xspace}
\usepackage{algorithm}
\usepackage{algpseudocode}
\usepackage{amssymb}
\usepackage{amsmath,amsfonts}
\usepackage[font=small,labelfont=bf]{caption}
\usepackage{multirow}
\usepackage{wrapfig}
\newcommand{\bA}{\mathbf{A}}

\newcommand{\bK}{\mathbf{K}}

\newcommand{\bM}{\mathbf{M}}

\newcommand{\bO}{\mathbf{O}}
\newcommand{\bp}{\mathbf{p}}\newcommand{\bP}{\mathbf{P}}
\newcommand{\bQ}{\mathbf{Q}}

\newcommand{\bS}{\mathbf{S}}

\newcommand{\bV}{\mathbf{V}}

\newcommand{\nR}{\mathbb{R}}

\newcommand{\cD}{\mathcal{D}}
\newcommand{\cE}{\mathcal{E}}

\newcommand{\cS}{\mathcal{S}}

\newcommand{\figref}[1]{Fig.~\ref{#1}}
\newcommand{\secref}[1]{Section~\ref{#1}}

\newcommand{\eqnref}[1]{Eq.~\eqref{#1}}

\newcommand{\tabref}[1]{Table~\ref{#1}}

\makeatletter
\DeclareRobustCommand\onedot{\futurelet\@let@token\@onedot}
\def\@onedot{\ifx\@let@token.\else.\null\fi\xspace}
\def\eg{e.g\onedot} 
\def\ie{i.e\onedot} 
 
\def\etc{etc\onedot}

\def\wrt{wrt\onedot}

\makeatother

\newcommand{\boldparagraph}[1]{\vspace{0.03cm}\noindent{\bf #1:}}

\definecolor{darkgreen}{rgb}{0,0.7,0}
\definecolor{darkblue}{RGB}{31,119,180}
\definecolor{darkred}{RGB}{214,39,40}

\usepackage{appendix}

\def\modelname{HDT}
\title{\modelname: Hierarchical Document Transformer}

\author{Haoyu He$^{1,2}$\thanks{Equal contribution.}, Markus Flicke$^{1,2 \ast}$, Jan Buchmann$^3$, Iryna Gurevych$^3$, Andreas Geiger$^{1,2}$ \\
$^1$ University of Tübingen  \qquad $^2$ Tübingen AI Center\\
$^3$ Technical University of Darmstadt and Hessian Center for AI (hessian.AI) \\
}

\colmfinalcopy

\begin{document}

\maketitle


\begin{abstract}
In this paper, we propose the Hierarchical Document Transformer (HDT), a novel sparse Transformer architecture tailored for structured hierarchical documents.
Such documents are extremely important in numerous domains, including science, law or medicine. However, most existing solutions are inefficient and fail to make use of the structure inherent to documents.
HDT exploits document structure by introducing auxiliary anchor tokens and redesigning the attention mechanism into a sparse multi-level hierarchy. This approach facilitates information exchange between tokens at different levels while maintaining sparsity, thereby enhancing computational and memory efficiency while exploiting the document structure as an inductive bias. We address the technical challenge of implementing HDT's sample-dependent hierarchical attention pattern by developing a novel sparse attention kernel that considers the hierarchical structure of documents. As demonstrated by our experiments, utilizing structural information present in documents leads to faster convergence, higher sample efficiency and better performance on downstream tasks.
\end{abstract}


\section{Introduction}

Many natural language processing tasks including summarization and question answering require language models to encode long structured documents such as scientific papers or Wikipedia articles into meaningful representations.
Since 2017, attention-based Transformer architectures~\cite{Vaswani2017NIPS} have established themselves as the dominant modeling paradigm, demonstrating state-of-the-art performance on numerous NLP tasks.
While dense attention yields context-rich representations, its computational complexity grows quadratically with the input.
This is problematic when processing large documents, in particular on resource constrained hardware such as consumer GPUs.

At the same time, most documents are naturally structured: Words form sentences, sentences form sections and sections form documents.
Surprisingly, this structure is largely ignored by many existing language models that typically consider their context as a ``flat'' sequence of tokens.
While transformer models can in principle learn to generalize hierarchically, \textit{structural grokking} requires extremely long training~\citep{Murty2023ACL}.
Moreover, the performance of flat long context models depends on the position of relevant information~\citep{Liu2023ARXIV}.
We hypothesize that exploiting the structure of documents explicitly yields two major benefits: (1) Imposing the structure of documents as an inductive bias improves sample efficiency and generalization. (2) Adapting the attention pattern to the document structure leads to sparse representations which reduce computational and memory complexity, and enable processing of long documents even on consumer hardware.

\begin{figure}[t!]
\centering
    \begin{subfigure}[ht]{0.68\textwidth}
        \includegraphics[width=\textwidth]{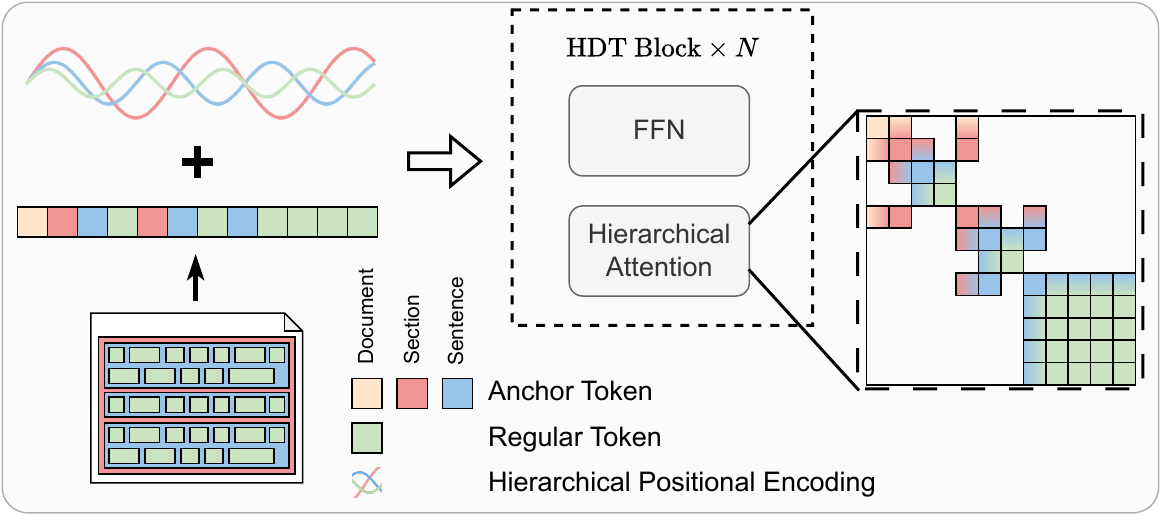}
        \caption{Model Architecture}
        \label{fig:model_architecture}   
    \end{subfigure}
    \hfill
    \begin{subfigure}[ht]{0.29\textwidth}
        \includegraphics[width=\textwidth]{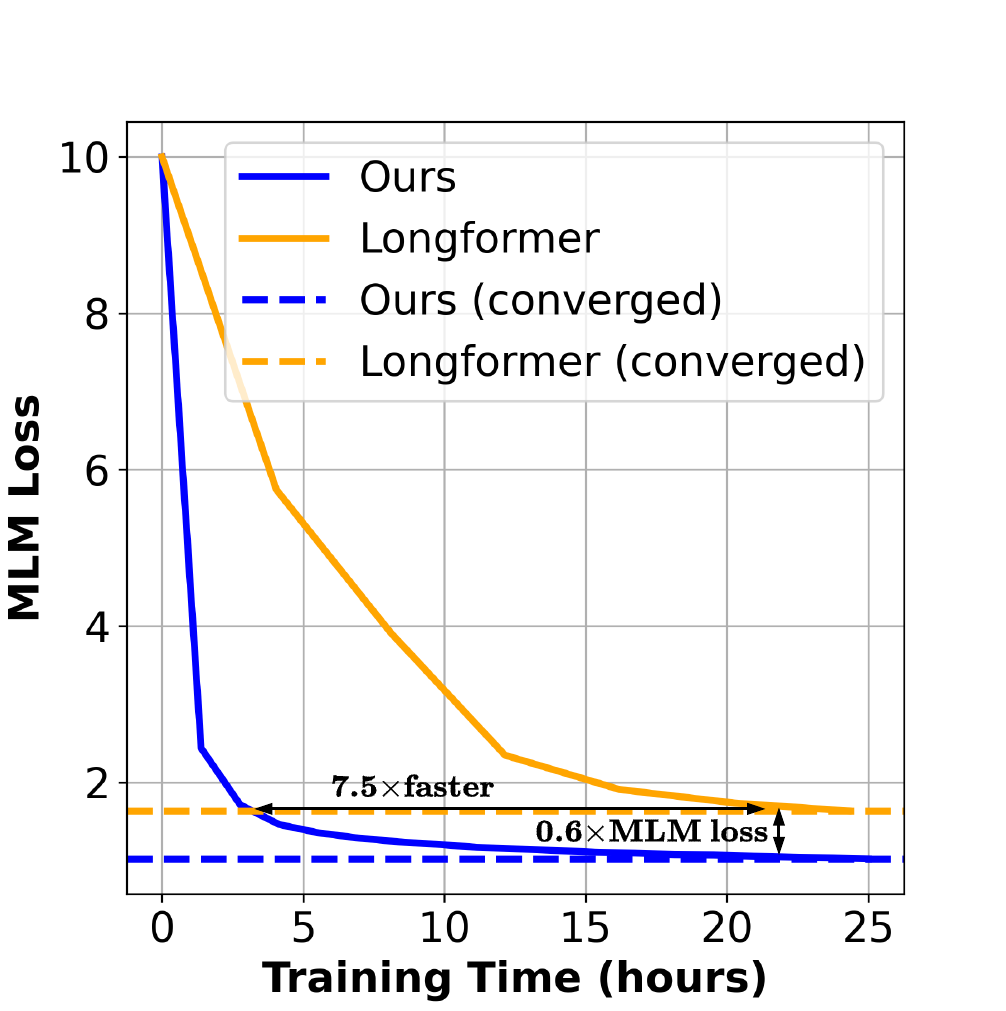}
        \caption{Convergence Speed}
        \label{fig:convergence}        
    \end{subfigure}
    \vspace{-0.1cm}
\caption{(\subref{fig:model_architecture}) We propose a sparse attention kernel that considers the hierarchical structure of documents. Here, regular tokens are illustrated in green, and auxiliary anchor tokens in yellow (document), red (section) and blue (sentence). Each token attends to its parent, siblings and children. Cross-level attention is illustrated using color gradients in the attention matrix. Utilizing structural information present in documents leads to faster pre-training (\subref{fig:convergence}) and better performance on downstream tasks. We use the held-out validation set in (\subref{fig:convergence}) to calculate the MLM loss.}
\vspace{-0.3cm}
\end{figure}

To test this hypothesis, we propose the \textit{Hierarchical Document Transformer (\modelname)}, a novel sparse Transformer architecture for processing hierarchically structured documents.
More specifically, we first introduce auxiliary anchor tokens for all structural elements such as sentences, sections and documents as illustrated in \figref{fig:model_architecture}.
Second, we redesign the attention pattern of a Transformer block into a multi-level hierarchy where information is exchanged only between tokens at the same level (siblings) as well as between the respective parent and child tokens.
By stacking multiple HDT blocks, information from any token can reach any other token.
At the same time, the attention pattern of HDT blocks is highly sparse, leading to gains in computational and memory efficiency.
In contrast to previous hierarchical models~\cite{Wu2021IJCNLP,Chalkidis2022ARXIV}, our HDT block supports deeper hierarchies and communicates at all hierarchy levels \textit{simultaneously} such that stacking of only a few HDT layers establishes communication between all tokens.

However, implementing our model required solving a key technical challenge: Existing sparse Transformer architectures like LongFormer \cite{Beltagy2020ARXIV} assume a fixed sparsity pattern for all inputs in a mini-batch which can be implemented using standard libraries. In contrast, HDT imposes a different sparsity pattern for each sample within a mini-batch as each document is structured differently. Towards this goal, we developed a novel, flexible and efficient attention kernel based on the Triton \citep{Tillet2019MAPL} library.
We pre-train HDT encoder models using Masked Language Modeling (MLM) and HDT encoder-decoder models using UL2~\citep{Tay2023ICLR} on three document datasets: arXiv~\citep{Saier2023JCDL}, HUPD Patents~\citep{Suzgun2023NEURIPSDATA} and Wikipedia.
We demonstrate improved pre-training convergence rates (see \figref{fig:convergence}) as well as better downstream task performance on several proximity, summarization, QA and NLI tasks.
%
Our code and data are available at \url{https://github.com/autonomousvision/hdt}.

\section{Related Work}

This study combines two lines of work that try to improve the performance of bi-directional Transformers in handling long-document inputs: \textit{structural modeling} and \textit{efficient attention}. 

Structural modeling approaches exploit the hierarchical organization of documents into sections, subsections, paragraphs, sentences, \etc. They can be divided into two categories \cite{Buchmann2024EACL}: (1) In \textit{structure infusion}, structural information is added to the Transformer input, \eg, via special tokens \citep{Aghajanyan2022ICLR, Liu2022ARXIVb, Buchmann2024EACL}, position embeddings \citep{Bai2021AAAI, Cao2022ACL}, attention masks \citep{Wang2019EMNLP, Liu2021EMNLP, Hong2022EMNLP, Zhang2021ACL} or fusing self-attention layers with GNN layers \cite{Sachan2020EACL, Ahmed2019ACL}. While observing performance improvements, these works do not
improve efficiency. (2) \textit{Hierarchical processing} employs architectures that first contextualize tokens on a local (\eg, sentence) level, aggregate the local representations (\eg, in a ``[CLS]'' token) and then contextualize the aggregates on one or several higher levels (\eg, paragraph or section, \citealt{Yang2016NAACL, Chalkidis2019ACL, Yang2019ARXIV, Ruan2022ACL, Zhang2022EMNLP, Dai2022EMNLP}). Local and higher-level contextualization is performed iteratively over several layers.
While hierarchical processing improves efficiency, separation of contextualization levels into different layers results in two problems: (1) Contextualization is not conducted simultaneously across all levels, hence more layers are required potentially. (2) Because different layers assume different functions, they can't serve as drop-in replacements for common Transformer architectures. HDT resolves these problems by simultaneous contextualization over all hierarchy levels in a \textit{single layer} using an efficient hierarchical attention pattern.

Due to the quadratic memory and time complexity of the self-attention block of Transformer models~\citep{Vaswani2017NIPS}, major efforts have been devoted to reducing complexity, in particular when applying attention to longer sequences.
One branch of work focuses on using sparse attention patterns to reduce computation while maintaining expressiveness
including Sparse Transformer \citep{Rewon2019ARXIV}, ETC \citep{Ainslie2020EMNLP}, BigBird \citep{Zaheer2020NEURIPS}, Longformer \citep{Beltagy2020ARXIV} and CoLT5 \citep{Ainslie2023EMNLP} among others. 
Another branch of work aims at speeding up attention computation by considering GPU hardware characteristics. FlashAttention 1 \& 2  \citep{Dao2022NEURIPS, Dao2023ARXIV} propose tiling with block-wise attention to reduce memory IOs and compute attention blocks in parallel in Static Random Access Memory (SRAM). \citet{Matteo2023ARXIV} extends the idea of FlashAttention to key/query dropping and hashing-based attention that was originally proposed by \citet{Kitaev2020ICLR}. Our work builds upon both branches: We propose a dynamic sample-dependent sparse attention pattern which exploits the structure of text documents. For efficiency, we follow \citet{Dao2022NEURIPS} and implement this pattern as a customized memory-aware kernel.   



\section{Methodology}

\begin{figure}[t!]
\includegraphics[width=\textwidth]{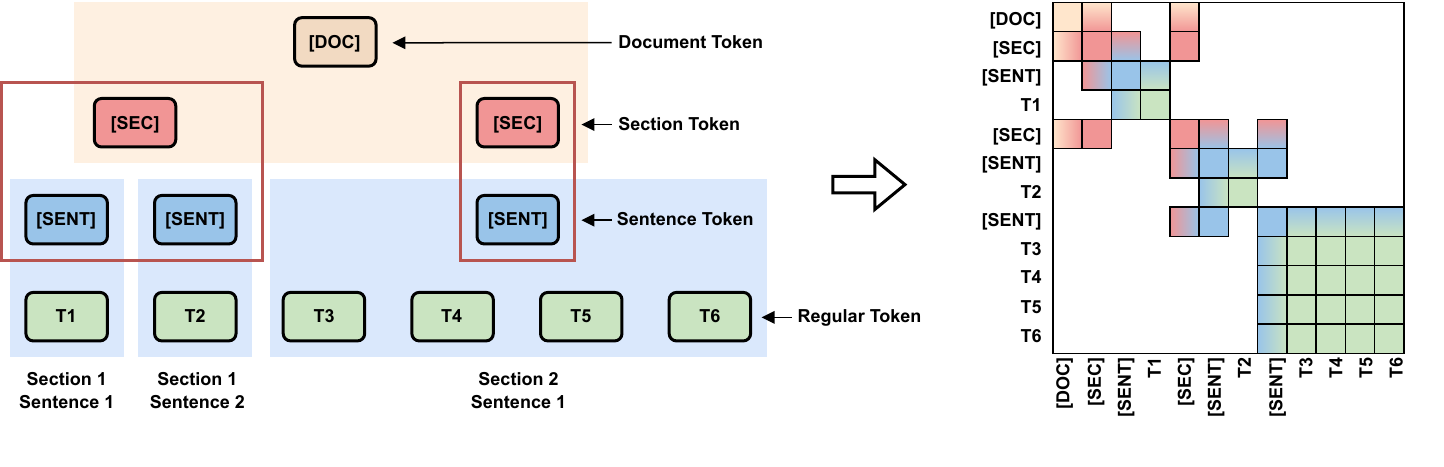}
\vspace{-0.8cm}
\caption{\textbf{Hierarchical Document Decomposition.} Left: Tree representation of a document.
Tokens within the same box attend to each other.
Tokens that do not share a box attend to each other only indirectly
(\eg, T1 and T3 via the sentence and section tokens). Right: Sparse attention matrix.}
\label{fig:model_hierarchy}
\vspace{-0.3cm}
\end{figure}


We now introduce the proposed Hierarchical Document Transformer (\modelname) for efficient long-document modeling. First, we briefly recap the standard transformer model. Next, we introduce the proposed \modelname~ block.
Finally, we describe the design of the encoder-only as well as encoder-decoder \modelname~architectures which are used in our experiments in \secref{sec:experiments}.

\subsection{Standard Transformer}

The original transformer encoder model by~\cite{Vaswani2017NIPS} is composed of Multi-Head Self-Attention (MHSA) layers that are interleaved with shallow feed-forward networks and residual connections. Because the attention pattern is identical across multiple heads in an attention layer, for simplicity of notation, we represent queries, keys and values for a single attention head as $\bQ,\bK,\bV \in \nR^{n\times d_k}$, where $n$ is the sequence length and $d_k$ the head dimension. All MHSA layers in the model, as well as the embedding
layers, produce outputs of dimension $d_{\text{model}}$. We further denote the number of heads as $h$, hence $d_{\text{model}}=d_k \times h$.  A self-attention head computes its output $\bO \in \nR^{n\times d_k}$ as follows:
\begin{equation}
    \bA=\frac{\bQ\bK^T}{\sqrt{d_k}} \qquad \bS=\text{softmax}(\bA) \qquad \bO=\bS\bV
    \label{eq:standard_transformer}
\end{equation}
where the softmax operator is applied row-wise.
The standard attention block has $O(n^2)$ time and memory complexity where $n$ is the input length. It further does not explicitly take the structure of documents into account.


\subsection{Hierarchical Document Transformer}
\label{sec:hdt}
Most documents are organized into structural constituents like sections, paragraphs, sentences, bulleted lists, figures, and footnotes. This structure is represented in the visual layout and conveys the author's semantic organization of the text
~\citep{Taylor1984RRQ, Guthrie1991RRQ}. 
While our model is general and can handle arbitrary hierarchical structures, for simplicity we will focus our exposition on a document hierarchy with three levels: tokens, sentences and sections.
More specifically, we split a document with $n$ tokens $\cD=(t_1, t_2, ..., t_n)$ into sections $\cD=(\cE_1, \cE_2, ..., \cE_{|\cD|})$. Sections $\cE_i$ are split into sentences $\cE_i=(\cS_1, \cS_2, ..., \cS_{|\cE_i|})$ which are split into sequences of regular tokens $\cS_j=(t_1, t_2, ..., t_{|\cS_j|})$.

We exploit this document structure by (1) introducing auxiliary anchor tokens to represent each element in the hierarchy, and (2) developing an efficient sparse attention kernel that exchanges information only between tokens at the same level (siblings) as well as between the respective parent and child tokens.
By stacking multiple HDT blocks, information from any token can reach any other token.
At the same time, the attention pattern of HDT blocks is highly sparse, leading to gains in computational (and memory) efficiency.
More specifically, for the document hierarchy introduced above, we prepend additional [SENT] anchor tokens to the beginning of every sentence, [SEC] anchor tokens to the start of each section, and a [DOC] anchor token to the beginning of the document as illustrated in \figref{fig:model_hierarchy}.



\begin{figure}[t!]
\begin{center}
\includegraphics[width=1\textwidth]{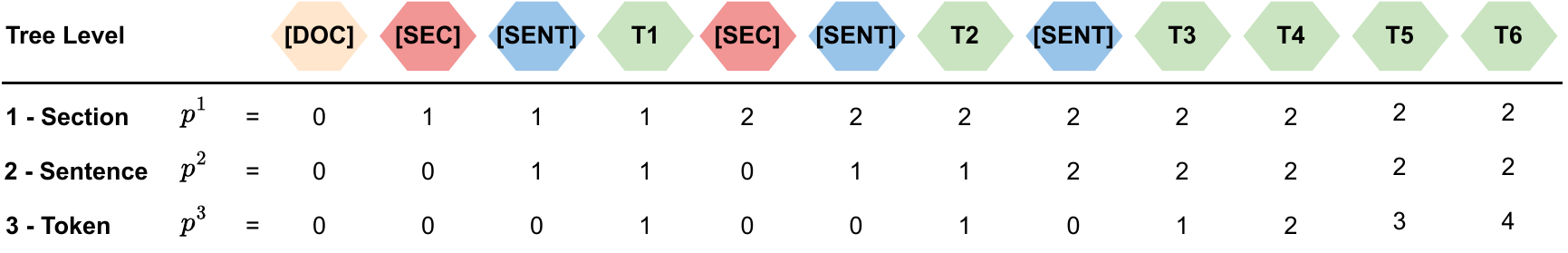}
\end{center}
\vspace{-0.2cm}
\caption{\textbf{Hierarchical Positional Encoding.}
We represent the position of each token in the hierarchy with one linear index $p^l$ per hierarchy level $l$ yielding an index vector $\mathbf{p}=(p^1,\dots,p^L)^T$. Above, we show an example with $L=3$ levels. Note that level 0 (document) does not require an index.
Each index in $\mathbf{p}$ is passed through sinusoidal encoding functions which are summed over all levels to form the final encoding vector according to \eqnref{eq:pos_enc}.}
\label{fig: indicies}
\vspace{-0.3cm}
\end{figure}

\boldparagraph{Hierarchical Positional Encoding}
We extend the sinusoidal position encoding to model $L$ hierarchy levels. To inform each token about its position within the hierarchy, we assign it one linear index $p^l$ per hierarchy level as illustrated in \figref{fig: indicies}, yielding an index vector $\mathbf{p}=(p^1,\dots,p^L)^T$. Each index in $\mathbf{p}$ is passed through a set of standard sinusoidal encoding functions which are summed over all levels to form the final hierarchical positional encoding (HPE) vector at $\bp$:
\begin{equation}
    {HPE}(\bp,i) = \sum_{l=1}^L
    \begin{cases}
      \sin(\omega_k\,p^l) & \text{if}\ i = 2k \\
      \cos(\omega_k\,p^l) & \text{if}\ i = 2k+1
    \end{cases}
    \quad \text{where} \quad
    \omega_k = \frac{1}{10000^{2k/d_{\text{model}}}}
    \label{eq:pos_enc}
\end{equation}

\begin{figure}[t!]
\begin{center}
\includegraphics[width=1\textwidth]{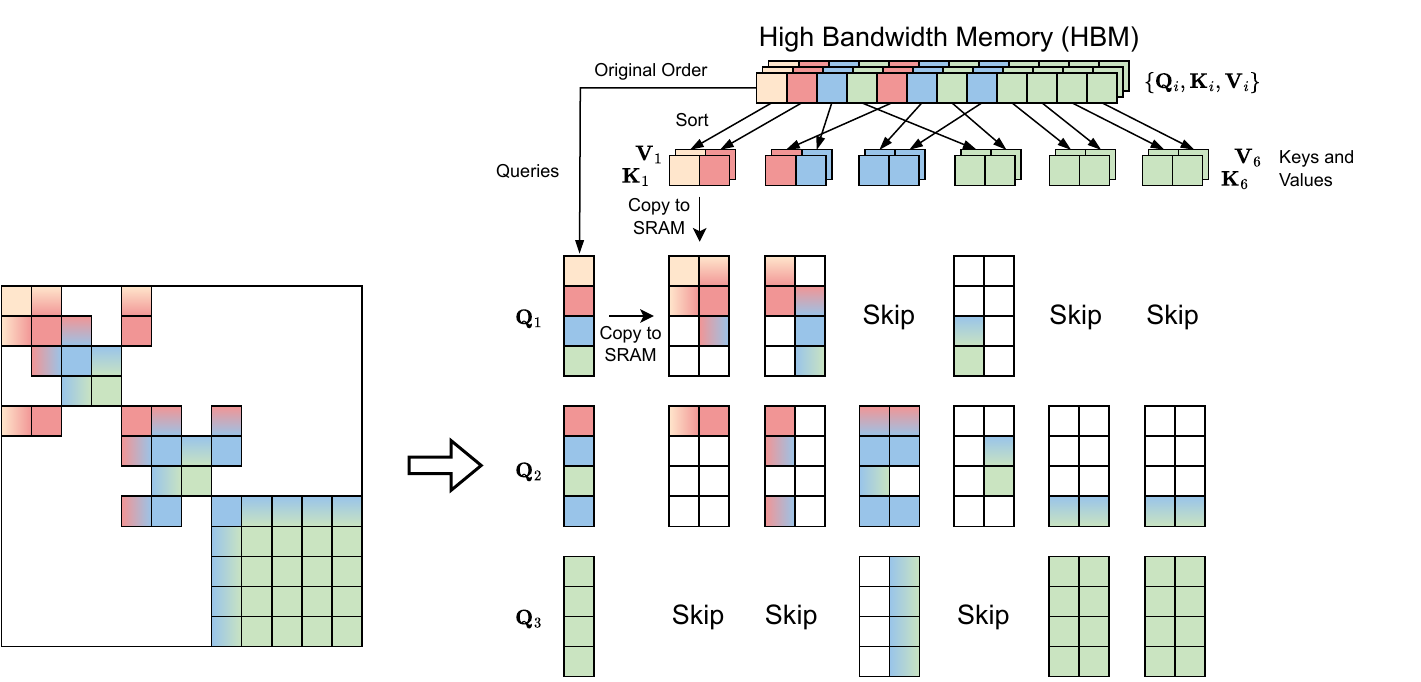}
\end{center}
\vspace{-0.2cm}
\caption{\textbf{Hierarchial Attention Kernel.}
We copy queries, keys and values block-wise to SRAM for fast attention computation using a fused kernel. To increase the number of empty blocks that can be skipped, we sort keys and values based on their hierarchy level. 
Larger examples are shown in \figref{fig:full_attn_mask}.
}
\label{fig:attn_flash}
\vspace{-0.3cm}
\end{figure}

\boldparagraph{Hierarchical Attention}
To model document structure, we impose a document-specific attention pattern by modifying the standard Transformer attention in \eqnref{eq:standard_transformer} as follows
\begin{equation}
    \bA=\frac{\bQ\bK^T}{\sqrt{d_k}} \qquad \bS=\text{softmax}(\bA \odot \mathbb{1}_\bM) \qquad \bO=\bS\bV
\end{equation}
with $(A_{ij} \odot \mathbb{1}_{M_{ij}})=A_{ij}$ if $M_{ij}=1$ and $-\infty$ if $M_{ij}=0$. For clarity of notation, we use $p^l_i, p^l_j$ to represent the position index for hierarchy level $l$ of the $i$'th token $t_i$ and the $j$'th token $t_j$, respectively.
The attention mask $\bM \in \{0, 1\}^{n \times n}$ is defined such that information is only exchanged directly between tokens at the same level as well as between the respective parent and child tokens.
For a 3-level document structure, we first compute the attention mask for each hierarchy level separately
\begin{align}
M_{ij}^{\text{DOC}} &= [p^2_i=0] \cdot [p^2_j=0]\\
M_{ij}^{\text{SEC}} &= [p^3_i=0] \cdot [p^3_j=0] \cdot [p^1_i=p^1_j]\\
M_{ij}^{\text{SENT}} &= [p^1_i=p^1_j] \cdot [p^2_i=p^2_j]
\end{align}
where $[\cdot]$ denotes the Iverson bracket which evaluates to 1 if the argument is true and 0 otherwise.
Finally, we perform an OR operation to obtain the full attention mask $\bM$:
\begin{equation}
    \bM = \bM^{\text{DOC}} \oplus \bM^{\text{SEC}} \oplus \bM^{\text{SENT}} 
    \label{eq:attn_mask}
\end{equation}
Note that $\bM$ is highly sparse in practice (see \figref{fig:full_attn_mask} for an example) and hence reduces theoretical complexity from $O(n^2)$ to $O(n\,s)$ where $s$ is the length of the longest sentence in the document. Thus, computational savings are largest for long documents for which $s \ll n$. However, when using parallel hardware (\eg, GPUs) as customary in deep learning, special care has to be taken regarding the kernel design to translate these theoretical savings into actual wall-clock time reduction. We hence develop a custom hierarchical attention kernel.

More specifically, we build upon the recent tiling-based ideas of FlashAttention~\citep{Dao2022NEURIPS, Dao2023ARXIV}. FlashAttention partitions attention computation into small $128\times64$ token \textit{blocks} that can be computed efficiently in SRAM. This is in contrast to classical attention implementations that materialize intermediate outputs to slow High Bandwidth Memory (HBM). However, na\"{i}vely partitioning the attention matrix into regular blocks is suboptimal given the uniformly distributed entries of $\bM$.
Furthermore, each document has a different structure and hence leads to a different sparsity pattern in the attention matrix.

To maximize the number of empty blocks that can be skipped, we leverage a simple heuristic which is illustrated with an example in \figref{fig:attn_flash}. Specifically, before copying keys and values to SRAM, we first sort them based on their hierarchy level from $l=0$ to $L$ while keeping the order of the queries unchanged. This ensures adjacency of the most related tokens and hence increases the probability of large empty blocks that can be skipped as illustrated in Appendix \figref{fig:full_attn_mask}. Afterwards, we copy the queries $\bQ_i$, keys $\bK_j$ and values $\bV_j$ of block $(i,j)$ to SRAM and apply block attention. We process all non-empty blocks in parallel, skipping empty ones. Finally, we write the result $\bO_i$ back to HBM using the online softmax algorithm~\citep{Milakov2018ARXIV}. The complete algorithm for a forward pass is provided in Appendix Algorithm~\ref{alg:hierarchical_attention_kernel}. 
While we found this simple sorting heuristic to work well in practice (see \figref{fig:kernel_runtime} for a runtime comparison to Block-Sparse FlashAttention and Appendix \figref{fig:full_attn_mask} for qualitative examples), more advanced permutation algorithms for sparse matrices \citep{Aykanat2004SICS,Ferris1998MP,Hendrickson2000SICS} could possibly lead to larger computational savings at the cost of additional inference time per data sample. We will leave the investigation of such trade-offs for future work.

\boldparagraph{Hierarchical Encoder and Decoder Stacks}
We realize the \textit{encoder} as a stack of $N$ identical \modelname~blocks where each block is composed of two sub-layers as illustrated in \figref{fig:model_architecture}. The first sub-layer performs hierarchical multi-head self-attention as described above, and the second is a simple position-wise fully connected feed-forward network (FFN). We use layer normalization at the beginning of each sub-layer as well as residual connections~\citep{He2016CVPR} for each layer. We apply the hierarchical positional encoding to the input tokens and add them to the respective token embeddings as input to the encoder stacks.
The \textit{decoder} is composed of a stack of $N$ identical standard Transformer decoder blocks, including a sub-layer of causal multi-head attention that prevents tokens from attending to subsequent positions, a sub-layer to perform multi-head cross attention \wrt the output of the encoder stack,  and a standard feed-forward network (FFN). Following T5 \citep{Raffel2020JMLR}, we use relative positional encoding for the decoder.

\section{Experiments}
\label{sec:experiments}


We first demonstrate the utility of structure-aware attention patterns on a simple mathematical task.
Next, we show the effectiveness of our encoder-only model on SciRepEval proximity tasks.
We also investigate the expressiveness of the anchor token representations using our encoder-decoder model on the FacetSum summarization tasks.
Experiments on SCROLLS demonstrate that our model can even be applied to long texts which are not explicitly structured.
Finally, we provide a detailed efficiency analysis.

\subsection{Mathematical Reasoning Tasks}

\begin{figure}[t!]
  \centering
  \begin{minipage}[b]{0.30\textwidth}
    \centering
    \includegraphics[width=\textwidth]{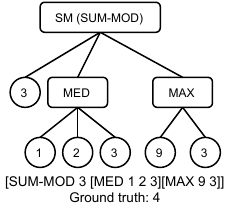}\\\vspace{0.4cm}
    \vspace{-0.6cm}
    \caption{ListOps Sample}
    \label{fig:listops_task}
  \end{minipage}
  \hfill
  \begin{minipage}[b]{0.30\textwidth}
    \centering
    \includegraphics[width=\textwidth]{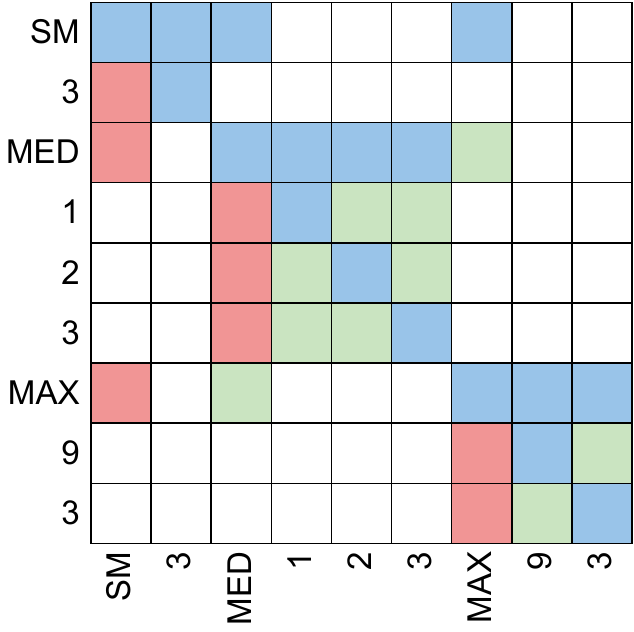} 
    \vspace{-0.6cm}
    \caption{ListOps Attention}
    \label{fig:red_green_blue} 
  \end{minipage}
  \hspace{0.2cm}
  \hfill
  \begin{minipage}[b]{0.34\textwidth}
    \centering
    \includegraphics[width=\textwidth]{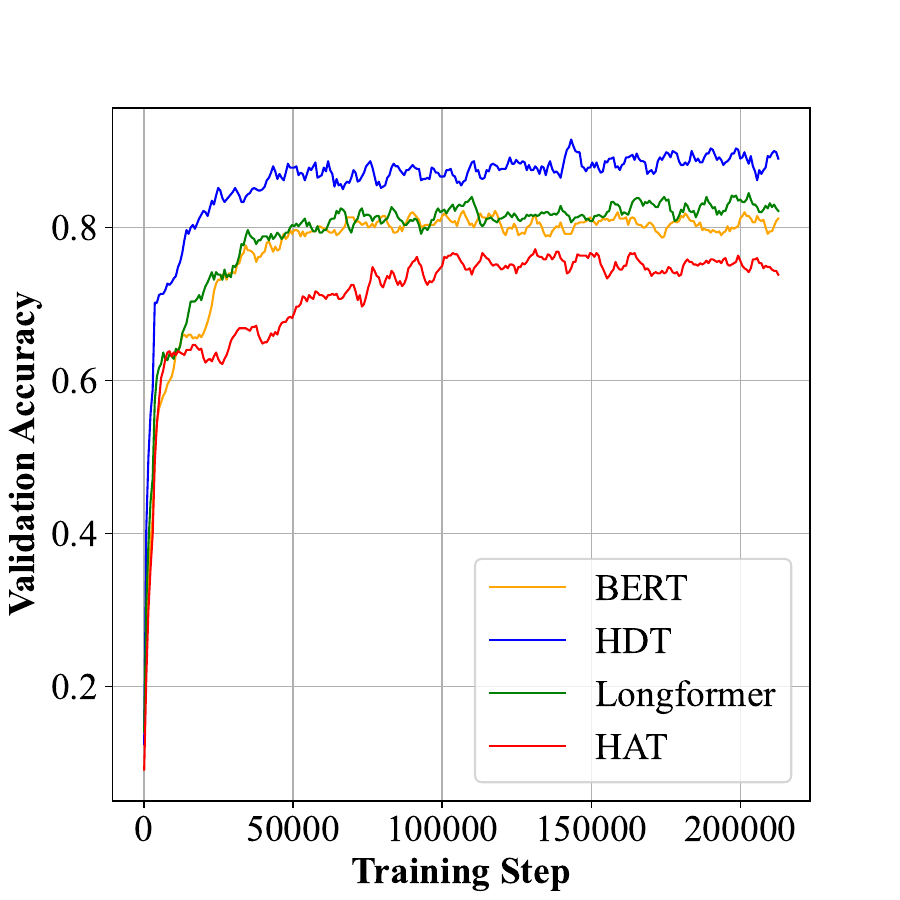} 
    \vspace{-0.63cm}
    \caption{ListOps Convergence}
    \label{fig:convergence_plots} 
  \end{minipage}
\vspace{-0.3cm}
\end{figure}

\begin{wraptable}{r}{0cm}
\begin{tabular}{l|c}
\toprule
\textbf{Model}  & \textbf{Acc.} \\ \hline
$\mathtt{BERT}$ & {75.9} \\ 
$\mathtt{\modelname}_\mathtt{r+g+b}$ &  {79.7} \\ 
$\mathtt{\modelname}_\mathtt{g+b}$ & {85.6}\\
$\mathtt{\modelname}_\mathtt{blue}$ & {\textbf{86.2}} \\ 
\bottomrule
\end{tabular}
\captionof{table}{ListOps Acc.}
\label{tab:listops}
\end{wraptable}

As proof of concept, we first compare our encoder model (\modelname-E) to BERT~\citep{Devlin2019NAACL}, Longformer~\citep{Beltagy2020ARXIV} and HAT~\citep{Chalkidis2022ARXIV} on the ListOps mathematical reasoning task~\citep{Nikita2018NAACL}. The ListOps dataset is composed of simple list operations (Modular Sum, Minimium, Maximum, Median) with 2-5 operands and maximum tree depth 20. An example is shown in \figref{fig:listops_task}. \figref{fig:red_green_blue} shows the different types of attention patterns we use. The combination of red+green+blue entries corresponds to the attention pattern defined in \secref{sec:hdt}. However, the special structure of the problem admits of further increasing sparsity by using a ``causal'' mask where operands do \textit{not} attend to their operators (green+blue) and where operands do \textit{not} attend to each other (blue only).
Our ablations in \tabref{tab:listops} show that as we increase the inductive sparsity bias, test set accuracy improves.
As illustrated in \figref{fig:convergence_plots}, whilst BERT, Longformer and HAT use a learned positional embedding, we find that HDT outperforms all baselines on ListOps even without using any positional embedding as the operators in the ListOps task are invariant to the positions of their operands.

\subsection{Language Tasks}
\label{sec:language_tasks}
We now investigate the utility of \modelname~for long-document processing tasks. As we are interested in efficiency, we conduct all experiments using a constrained compute budget. Following \citet{Geiping2023ICML}, we pre-train our encoder models (\modelname-E) for 24 hours on 1 GPU ($\sim$10k steps) and our encoder-decoder models (\modelname-ED) for 72 hours on 4 GPUs ($\sim$50k steps) using a batch size of 128.
Compared to the baselines, our models require 5-20 times fewer pre-training steps to reach better or comparable downstream task performance.
We leave scaling \modelname~to larger sizes and more data for future work.

\boldparagraph{Data}
To train our proposed structure-aware models, we build a large full-text document corpus from unarXive \citep{Saier2023JCDL} (1.9M arXiv papers), HUPD \citep{Suzgun2023NEURIPSDATA} (utility patent applications) and the latest Wikipedia dump processed with Gensim~\citep{Rehurek2010LRECWORK}. In total, our corpus includes over 12M long documents with extracted structural information. Our word tokenizer utilizes a vocabulary size of 32,768 tokens, trained with Byte Pair Encoding (BPE)~\citep{Sennrich2016ACL}.
For splitting sentences, we use the NLTK sentence tokenizer\footnote{https://www.nltk.org/api/nltk.tokenize.html}. 
We report results of \modelname-E pre-trained with and without arXiv data on a full-document version of SciRepEval~\citep{Singh2023EMNLP}. We evaluate \modelname-ED on the FacetSum~\citep{Meng2021ACL} and SCROLLS~\citep{Shaham2022EMNLP} benchmarks which contain tasks for long text reasoning requiring context modeling.
Benchmarks details are provided in Appendix \secref{sec:benchmarks}.


\boldparagraph{Baselines}
We compare \modelname-E to Hierarchical Attention Transformer (HAT)~\citep{Chalkidis2022ARXIV} which is the closest work to us that applies attention hierarchically to reduce computation/memory usage for processing long text. \modelname-E differs from HAT as follows: (1) \modelname-E takes the natural structure of text to build hierarchies while HAT cuts long text into fixed-size segmentations that do not adheare to the structure of natural language, (2) \modelname-E enables simultaneous information passing across all levels in a single layer while HAT requires multiple blocks to contextualize information between hierarchies. Moreover, HAT only considers two hierarchy levels. Despite the flexibility of our model, it achieves comparable latency/throughput supported due to our highly efficient kernel implementation.
As the sparse attention baseline, we choose Longformer~\citep{Beltagy2020ARXIV} and its encoder-decoder variant Longformer-Encoder-Decoder (LED) which are effective on various long-text tasks in previous works \citep{Tay2021ICLR, Dasigi2021NAACL, Tay2023CS}. We also include the current SotA methods SciBERT \citep{Beltagy2019EMNLP} and SciNCL \citep{Ostendorff2022EMNLP} as additional SciRepEval baselines. Both baselines use dense attention with an input length of 512 capturing only title and abstract. Unless stated otherwise, we use the officially released code and models by the original authors.

\boldparagraph{Pre-training \& Fine-tuning}
%
We pre-train the encoder-only model \modelname-E and the encoder-decoder model \modelname-ED from scratch on our training corpus. All models are pre-trained using an input length of 8,192 tokens and a mini-batch size of 128 (via gradient accumulation).
\modelname-E is pre-trained on the standard Masked Language Modeling (MLM) objective with a mask ratio of 15\%.
We also pre-train Longformer from scratch in the same setting as ours to study the effectiveness of the attention pattern we propose. \figref{fig:convergence} shows that our model converges faster than Longformer. \modelname-ED is pre-trained on UL2 \citep{Tay2023ICLR}, which is a unified pre-training paradigm with a range of denoising tasks.
We fine-tune all models on the downstream tasks using the default settings, see \secref{sec:settings} for details.

\begin{table}[t!]
\centering
\setlength{\tabcolsep}{4pt}
\begin{tabular}{lc|cccc|cc|c}
\toprule
\multirow{2}{*}{\textbf{Model}} & \textbf{Full} & \multicolumn{4}{c|}{\textbf{SciDocs}} & \multirow{2}{*}{\textbf{Feeds-M}} &	\textbf{High.} & \multirow{2}{*}{\textbf{Avg.}} \\
& \textbf{Text}  &   \textbf{Cite}     & \textbf{CoCite} & \textbf{CoView}  & \textbf{CoRead} & &  \textbf{Infl.} &       \\ \midrule
\multicolumn{9}{l}{\textbf{Pretrained Only}} \\ \midrule
$\mathtt{SciBERT}_{\mathtt{base}}$ &   & 53.75 &	66.73 &	66.37 &	53.20 &	63.18 &	40.80 &  57.34  \\
$\mathtt{Longformer}$ & \checkmark  & 56.64 &	71.92 &	71.51 &	61.57  &	63.66 &	43.82 &   61.52  \\ 
$\mathtt{HAT}$ & \checkmark  & 60.14  & 75.55  &	73.62 & 67.65 &	65.02 & \textbf{45.81}	 &  64.63  \\ 
$\mathtt{\modelname}$-$\mathtt{E}$ &  \checkmark & \textbf{62.50} &	\textbf{78.51} &	\textbf{75.69} &	\textbf{72.12}  &	65.19 &	43.60 &  \textbf{66.27}   \\ 
$\mathtt{\modelname}$-$\mathtt{E}$ (-$\mathtt{arXiv}$) &  \checkmark & 59.03 &	76.05 &	72.85 &	71.71  & \textbf{65.41} &	43.69 &  64.79   \\ \midrule
\multicolumn{9}{l}{\textbf{Pretrained + Finetuned with Contrastive Learning}} \\ \midrule
$\mathtt{SciNCL}$ @684k &  & 64.77 &	81.67 &	78.55 &	77.48  &	70.22 &	48.66 & 70.23    \\ 
\hline
$\mathtt{SciNCL}$ @19k &  & 62.56 &	82.29 &	77.84 &	75.84  &	67.11 &	46.23 & 68.65    \\ 
$\mathtt{Longformer}$ @19k & \checkmark  & 61.75  &	79.87 &	78.20 &	74.25 &	67.80 &	43.85 &	  67.62    \\ 
$\mathtt{HAT}$ @19k & \checkmark  & 63.46  & 81.24  &	\textbf{79.43} & 75.76  &	69.31 & 47.37	 &  69.42  \\ 
$\mathtt{\modelname}$-$\mathtt{E}$ @19k  & \checkmark  & \textbf{64.23} &	\textbf{82.44} & 78.95 & \textbf{77.09} &	\textbf{71.22} & \textbf{49.37} &  \textbf{70.55}   \\
$\mathtt{\modelname}$-$\mathtt{E}$ @19k (-$\mathtt{arXiv}$) & \checkmark  & 63.34 &	82.18 & 79.06 & 76.78 &	70.64 & 48.95 &  70.16   \\
\bottomrule
\end{tabular}
\caption{
\textbf{Results on SciRepEval Proximity Tasks.}
Top: Models pre-trained with MLM without fine-tuning. Bottom: Models pre-trained with MLM and fine-tuned using SciNCL's contrastive learning objective. Full text documents are available only for a subset of 19k training triplets. For reference, we also report the results of the original SciNCL model which is trained on all 684k title+abstract triplets. We also report $\mathtt{\modelname}$-$\mathtt{E}$ pre-trained without arXiv data (-$\mathtt{arXiv}$) to study the impact of scientific documents as pre-training data to our model's performance on the SciRepEval tasks which are in the scientific domain. All numbers are mean average precision. $\mathtt{SciBERT}$ and $\mathtt{SciNCL}$ use only title and abstract as input. }
\label{tab:scirepeval}
\vspace{-0.5cm}
\end{table}


\boldparagraph{Results on SciRepEval (Encoder Models)}
SciRepEval~\citep{Singh2023EMNLP} is a scientific document representation benchmark containing various classification, regression, and proximity tasks. Unfortunately, the original SciRepEval dataset comprises only titles and abstracts as for many papers in the original benchmark the full text is not publicly available. To adapt SciRepEval for long document models, we hence consider a subset of SciRepEval for which full-text articles are available in unarXive \citep{Saier2023JCDL}. \tabref{tab:scirepeval} shows the tasks with the largest number of samples, which are proximity tasks involving ranking a set of candidate papers by their relatedness to a query paper. Results on other SciRepEval tasks can be found in Appendix \tabref{tab:reg}. Without any fine-tuning, we take the hidden representation of the first token (either [CLS] for baseline models or [DOC] for \modelname-E) and sort papers by cosine similarity between paper representations. The first set of results in \tabref{tab:scirepeval} (``Pretrained Only'') shows the mean average precision for models pre-trained on MLM. Our results demonstrate that both, pre-training on full-text documents and considering the structure of documents, leads to substantial performance gains in this setting. Moreover, our model pre-trains significantly faster (10k iterations) than Longformer (64k) and HAT (50k) and can be trained from scratch while Longformer and HAT require parameter initialization from BART~\citep{Lewis2022ACL} and RoBERTa~\citep{Liu2019ARXIV1} (respectively) for best performance. This result underscores the sample efficiency of our structure-aware \modelname~model. 


Currently, the state-of-the-art performance on SciRepEval is reported by SciNCL~\citep{Ostendorff2022EMNLP} which finetunes a pre-trained SciBERT~\citep{Beltagy2019EMNLP} model using a contrastive learning objective. When comparing in this setting (``Pretrained + Finetuned with Contrastive Learning''), we observe that \modelname-E again outperforms all baselines, despite with a smaller margin. Notably, \modelname-E outperforms SciNCL even though SciNCL has access to 684k triplets for finetuning while \modelname-E is trained only on those 19k triplets for which full-text documents are publicly available. In addition, to test our model's reliance on scientific documents for pre-training, we \textit{exclude} the entire arXiv corpus from our pre-training dataset. We observed that our model's performance nearly matches HAT in the pre-training-only setting and surpasses HAT in the contrastive learning setting, even when using only out-of-domain data for pre-training.

\boldparagraph{Results on FacetSum (Encoder-Decoder Models)}
A major advantage of our model is its flexibility. As tokens are hierarchically grouped via anchor tokens, the document's content is compressed at different levels of granularity. We can query this information by letting the decoder of \modelname-ED attends to only a subset of the tokens or anchor tokens (\eg, [SEC]).
%
%
To better understand how much information is retained where in our model, we design a summarization task based on FacetSum~\citep{Meng2021ACL} which provides section-level summaries for Emerald journal articles. 
\tabref{tab:facetsum} shows our results. We observe performance on par with $\mathtt{LED}_\mathtt{base}$ even when attending only to the anchor tokens, which demonstrates the expressiveness of the learned intermediate representations. When additionally attending to all regular tokens (as $\mathtt{LED}_\mathtt{base}$ does), performance increases further. The right part of the table shows generalization performance when training on ``Findings'' and ``Value'', but evaluating ``Purpose'' and ``Method'' summaries.

\begin{table}[!tb]
\centering
\setlength{\tabcolsep}{5pt}
\begin{tabular}{l|cccc|cc}
\toprule
\textbf{Model}  & \textbf{Purpose} & \textbf{Method} & \textbf{Findings} & \textbf{Value} & \textbf{Purpose-ZS} & \textbf{Method-ZS} \\
\hline
$\mathtt{LED}_\mathtt{base}$ & 39.51 &  19.31 & 19.22  & \textbf{24.26} & 18.54 & 14.12 \\
$\mathtt{\modelname}$-$\mathtt{ED}$-$\mathtt{[SEC]}$ & 30.70  & 17.65  & 17.15 & 18.90 & 15.01 & 12.67 \\ 
+$\mathtt{[SENT]}$ & 34.29 & 19.72 & 18.38 & 20.53 & 19.67 & 13.94 \\
+ $\mathtt{tokens}$ & \textbf{40.60} & \textbf{22.21} & \textbf{22.11} & 22.11 & \textbf{21.75} & \textbf{15.43} \\
\bottomrule
\end{tabular}
\caption{\textbf{Results on FacetSum Summarization Task.} Following the original paper, we report ROUGE-L as the metric here. For $\mathtt{\modelname}$-$\mathtt{ED}$-$\mathtt{[SEC]}$, the decoder cross-attends only to the section anchor token [SEC].  We observe that even when attending only to the anchor tokens (+$\mathtt{[SENT]}$), our model is on par with LED, where the decoder attends to all tokens of the section, demonstrating the expressiveness of the learned intermediate representation of anchor tokens. When attending to additional regular tokens (+$\mathtt{tokens}$), our model outperforms LED. We also report zero-shot (ZS) performance for ``Purpose'' and ``Method'', training only on ``Findings'' and ``Value''.
}
\label{tab:facetsum}
\vspace{-0.45cm}
\end{table}

\boldparagraph{Results on SCROLLS (Encoder-Decoder Models)}
While our model benefits from the structure of documents as present in our SciRepEval and FacetSum evaluations, we are also interested in the applicability of our model to less structured ``flat'' long-text tasks.
Towards this goal, we evaluate our model on the SCROLLS benchmark, which contains 7 tasks spanning multiple domains that require reasoning over long texts for which document structure is not available. To apply our model to such inputs, we use pseudo sections composed of a fixed number of 32 sentences for hierarchical attention, while sentences are extracted as usual.
We fine-tune \modelname-ED on the 7 tasks in SCROLLS separately and submit the test set predictions to the public leaderboard\protect\footnote{\url{https://www.scrolls-benchmark.com/leaderboard}}.
\tabref{tab:scrolls} show our results. Compared to $\mathtt{LED}_{\mathtt{base}}$ with a similar number of parameters, \modelname-ED improves by over 2 score points. This demonstrates that our model is also effective on ``flat'' long-text tasks. However, unsurprisingly, our results are not competitive with state-of-the-art billion-parameter models such as CoLT5 XL which are trained for many epochs on large industrial GPU clusters.

\begin{table}[t!]
\centering
\resizebox{1.0 \linewidth}{!}{%
\setlength{\tabcolsep}{3pt}
\begin{tabular}{l|ccccccc|c}
\toprule
\multirow{2}{*}{\textbf{Model}} &  \textbf{GovRep} & \textbf{SumScr} &  \textbf{QMSum}  & \textbf{Qspr} & \textbf{Nrtv} & \textbf{QALT} & \textbf{CNLI} & \textbf{Avg} \\
           &   ROUGE-1/2/L     & ROUGE-1/2/L & ROUGE-1/2/L  & F1      & F1 & EM-T/H  &  EM & Score      \\ \hline
$\mathtt{LED}_{\mathtt{base}}$ & \textbf{56.2}/\textbf{26.6}/\textbf{28.8}   & 24.2/4.5/15.4
 & 25.1/\textbf{6.7}/\textbf{18.8} &	26.6 &	\textbf{18.5} &	25.8/25.4 & 71.5 &  29.16 \\
$\mathtt{\modelname}$-$\mathtt{ED}$ & 49.8/22.2/25.8  & \textbf{30.8}/\textbf{7.1}/\textbf{18.6} & \textbf{28.3}/\textbf{6.7}/18.7 & \textbf{33.1} & 14.2 & \textbf{29.4}/\textbf{26.4} & \textbf{81.4} & \textbf{31.41} \\ 
\bottomrule
\end{tabular}}
\caption{\textbf{Results on the SCROLLS Summarization, QA and NLI Benchmark.} We compare HDT-ED (pre-trained for 12 GPU days) to Longformer-Encoder-Decoder (LED) on the official SCROLLS benchmark \textit{without document structure}. We choose LED as baseline as it has a comparable number of parameters (162M) to HDT-ED (124M). We remark that neither model is competitive with state-of-the-art billion-parameter models such as CoLT5 XL (score 43.5) which are trained on large GPU clusters.}

\label{tab:scrolls}
\vspace{-0.3cm}
\end{table}

\boldparagraph{Efficiency Analysis}
Denoting the length (number of tokens) of the longest sentence in a document as $s$, the theoretical complexity of \modelname~attention is $O(n \times s)$. 
\figref{fig:kernel_runtime} compares GPU runtime and memory usage of different attention layers, including standard dense attention, block-sparse FlashAttention (using our pattern), Longformer sparse windowed attention, and \modelname~attention. 
All kernels are fed with real data, \ie, documents, which are further transformed to multi-head queries, keys, and values using 12 heads with head dimension 64. We report the runtime for one batch with 4 samples on an NVIDIA A100 GPU, and plot peak GPU memory consumption at lengths 16k and 32k for each kernel, except standard attention which exceeds 40 GB at 16k tokens.
Besides, we also compare runtime and memory consumption for three complete 12-layer models using different attention mechanisms in \tabref{tab:complexity}. Despite its flexibility, \modelname-E achieves runtime and memory consumption on par with HAT which uses 2 layers and fixed segment length.


\begin{minipage}{\textwidth}
  \begin{minipage}[b]{0.4\textwidth}
    \centering
    \includegraphics[width=\textwidth]{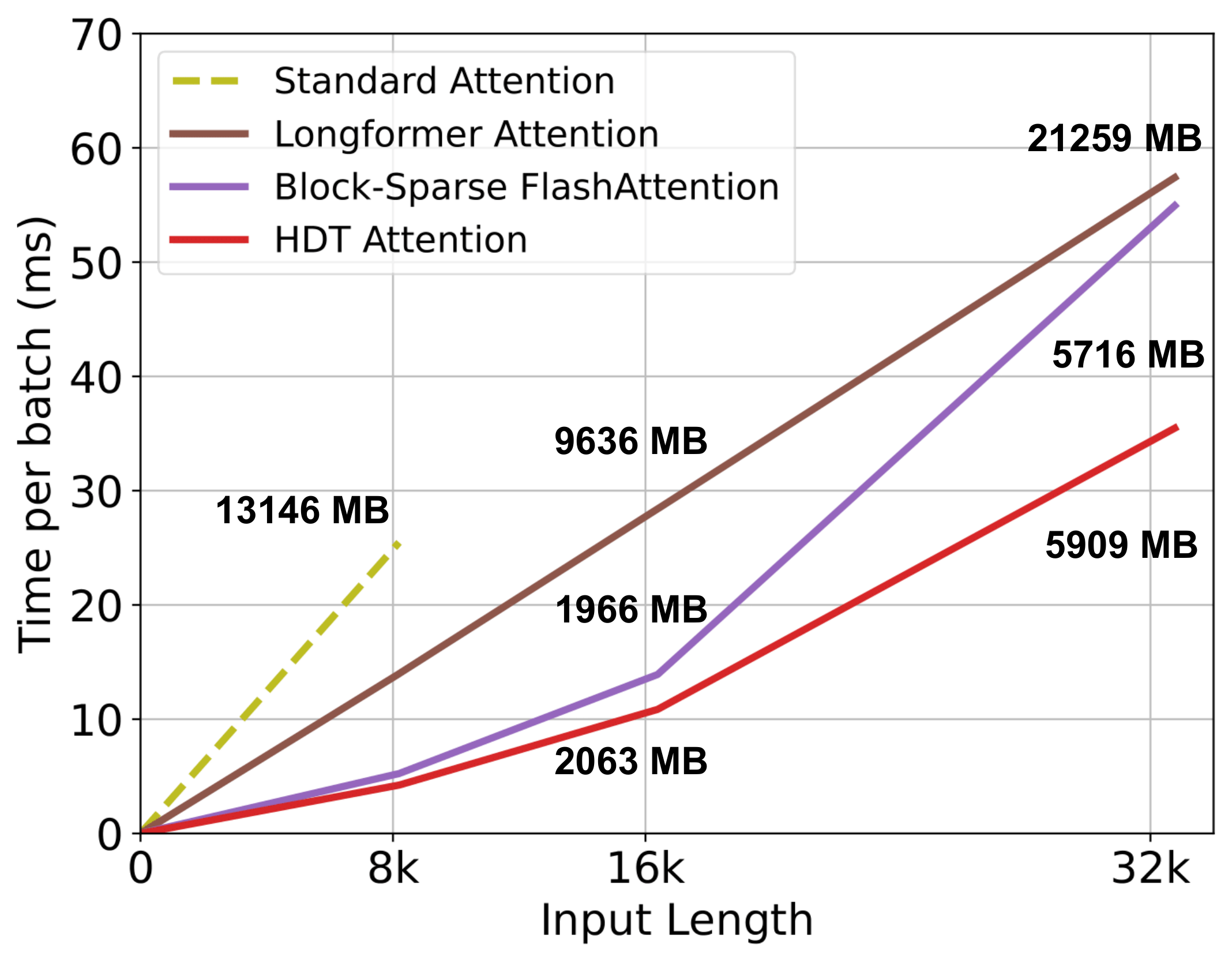}
    \vspace{-0.5cm}
    \captionof{figure}{
    \textbf{Runtime and Memory} consumption of a \textit{single attention layer}.
    }
    \label{fig:kernel_runtime}
  \end{minipage}
  \hfill
  \begin{minipage}[b]{0.55\textwidth}
    \centering
    \setlength{\tabcolsep}{4pt}
    \begin{tabular}{l|ccc}\hline
      \textbf{Model} & 
      $\mathtt{Longformer}$ & $\mathtt{HAT}$ & $\mathtt{\modelname}$-$\mathtt{E}$ \\ \hline
        \textbf{Complex.} & $O(n\,w)$ & $O(n\,k)$ & $O(n\,s)$ \\
        \textbf{\#Params} &  148.66 M  & 152.73 M & \textbf{108.99 M} \\
        \textbf{Time (ms)} & 178.82 \scriptsize{$\pm$ 6.84} & \textbf{77.84} \scriptsize{$\pm$ 2.30} & 79.8 \scriptsize{$\pm$ 2.96}  \\
        \textbf{TFLOPS} & 5.29 \scriptsize{$\pm$ 0.19} & 8.95 \scriptsize{$\pm$ 0.26} & \textbf{8.99 \scriptsize{$\pm$ 0.34}} \\ 
        \textbf{Memory} & 11.25 GB & \textbf{5.3 GB} & 5.85 GB \\ 
        \hline
      \end{tabular}
      \captionof{table}{\textbf{Runtime and Memory} consumption of several \textit{long-document models} with 12 layers. We report complexity, parameters, inference time, throughput, and memory usage using context length $n=4096$ and mini-batch size 1. Here, $w=512$ is the Longformer window size, $k=128$ is the fixed HAT segment length. $s$ is the length of the longest sentence in the document. 
      }
  \label{tab:complexity}
\end{minipage}
\vspace{-0.2cm}
\end{minipage}

\section{Conclusion}
We presented Hierarchical Document Transformer (\modelname), a novel approach for encoding long documents efficiently. By explicitly incorporating document structure into the attention mechanism, we achieve sparse representations, reducing computational complexity while improving sample efficiency and generalization.
We believe that hierarchical text representations offer many exciting opportunities in the future:
Extending the hierarchical structure down to byte-level could enable token-free language models.
Hierarchical ideas might also inspire novel decoder architectures that 
generate language in a structured hierarchical fashion.
Hierarchical models may also be advantageously combined with other models such as state space models, RNNs or ConvNets.
Finally, it still remains unclear if scaling laws also hold for hierarchical language models as they do for LLMs, and if hierarchical language models hold similar potential for unlocking emergent abilities.


\subsubsection*{\textbf{Acknowledgments}}
Andreas Geiger is a member of the Machine Learning Cluster of Excellence, funded by the Deutsche Forschungsgemeinschaft (DFG, German Research Foundation) under Germany’s Excellence Strategy – EXC number 2064/1 – Project number 390727645.
Jan Buchmann was funded by the European Union (ERC, InterText, 101054961). Views and opinions expressed are however those of the author(s) only and do not necessarily reflect those of the European Union or the European Research Council. Neither the European Union nor the granting authority can be held responsible for them.

\clearpage
\bibliography{bib/bibliography_long, bib/bibliography, bib/bibliography_custom}

\begin{thebibliography}{68}
\providecommand{\natexlab}[1]{#1}
\providecommand{\url}[1]{\texttt{#1}}
\expandafter\ifx\csname urlstyle\endcsname\relax
  \providecommand{\doi}[1]{doi: #1}\else
  \providecommand{\doi}{doi: \begingroup \urlstyle{rm}\Url}\fi

\bibitem[Aghajanyan et~al.(2022)Aghajanyan, Okhonko, Lewis, Joshi, Xu, Ghosh, and Zettlemoyer]{Aghajanyan2022ICLR}
Armen Aghajanyan, Dmytro Okhonko, Mike Lewis, Mandar Joshi, Hu~Xu, Gargi Ghosh, and Luke Zettlemoyer.
\newblock {HTLM}: Hyper-text pre-training and prompting of language models.
\newblock In \emph{Proc. of the International Conf. on Learning Representations (ICLR)}, 2022.

\bibitem[Ahmed et~al.(2019)Ahmed, Samee, and Mercer]{Ahmed2019ACL}
Mahtab Ahmed, Muhammad~Rifayat Samee, and Robert~E. Mercer.
\newblock You only need attention to traverse trees.
\newblock In \emph{Annual Meeting of the Association for Computational Linguistics (ACL)}, 2019.

\bibitem[Ainslie et~al.(2020)Ainslie, Onta{\~{n}}{\'{o}}n, Alberti, Cvicek, Fisher, Pham, Ravula, Sanghai, Wang, and Yang]{Ainslie2020EMNLP}
Joshua Ainslie, Santiago Onta{\~{n}}{\'{o}}n, Chris Alberti, Vaclav Cvicek, Zachary Fisher, Philip Pham, Anirudh Ravula, Sumit Sanghai, Qifan Wang, and Li~Yang.
\newblock {ETC:} encoding long and structured inputs in transformers.
\newblock In \emph{Conference on Empirical Methods in Natural Language Processing (EMNLP)}, 2020.

\bibitem[Ainslie et~al.(2023)Ainslie, Lei, de~Jong, Onta{\~{n}}{\'{o}}n, Brahma, Zemlyanskiy, Uthus, Guo, Lee{-}Thorp, Tay, Sung, and Sanghai]{Ainslie2023EMNLP}
Joshua Ainslie, Tao Lei, Michiel de~Jong, Santiago Onta{\~{n}}{\'{o}}n, Siddhartha Brahma, Yury Zemlyanskiy, David~C. Uthus, Mandy Guo, James Lee{-}Thorp, Yi~Tay, Yun{-}Hsuan Sung, and Sumit Sanghai.
\newblock Colt5: Faster long-range transformers with conditional computation.
\newblock In \emph{Conference on Empirical Methods in Natural Language Processing (EMNLP)}, 2023.

\bibitem[Aykanat et~al.(2004)Aykanat, Pinar, and {\c{C}}ataly{\"{u}}rek]{Aykanat2004SICS}
Cevdet Aykanat, Ali Pinar, and {\"{U}}mit~V. {\c{C}}ataly{\"{u}}rek.
\newblock Permuting sparse rectangular matrices into block-diagonal form.
\newblock \emph{SIAM Journal on Scientific Computing (SICS)}, 25\penalty0 (6):\penalty0 1860--1879, 2004.

\bibitem[Bai et~al.(2021)Bai, Shi, Lin, Xie, Tan, Xiong, Gao, and Li]{Bai2021AAAI}
He~Bai, Peng Shi, Jimmy Lin, Yuqing Xie, Luchen Tan, Kun Xiong, Wen Gao, and Ming Li.
\newblock Segatron: Segment-aware transformer for language modeling and understanding.
\newblock In \emph{Proc. of the Conf. on Artificial Intelligence (AAAI)}, 2021.

\bibitem[Beltagy et~al.(2019)Beltagy, Lo, and Cohan]{Beltagy2019EMNLP}
Iz~Beltagy, Kyle Lo, and Arman Cohan.
\newblock Scibert: {A} pretrained language model for scientific text.
\newblock In \emph{Conference on Empirical Methods in Natural Language Processing (EMNLP)}, 2019.

\bibitem[Beltagy et~al.(2020)Beltagy, Peters, and Cohan]{Beltagy2020ARXIV}
Iz~Beltagy, Matthew~E. Peters, and Arman Cohan.
\newblock Longformer: The long-document transformer.
\newblock \emph{arXiv.org}, 2004.05150, 2020.

\bibitem[Buchmann et~al.(2024)Buchmann, Eichler, Bodensohn, Kuznetsov, and Gurevych]{Buchmann2024EACL}
Jan Buchmann, Max Eichler, Jan-Micha Bodensohn, Ilia Kuznetsov, and Iryna Gurevych.
\newblock Document structure in long document transformers.
\newblock In \emph{Conference of the European Chapter of the Association for Computational Linguistics (EACL)}, 2024.

\bibitem[Cao \& Wang(2022)Cao and Wang]{Cao2022ACL}
Shuyang Cao and Lu~Wang.
\newblock {HIBRIDS:} attention with hierarchical biases for structure-aware long document summarization.
\newblock In \emph{Annual Meeting of the Association for Computational Linguistics (ACL)}, 2022.

\bibitem[Chalkidis et~al.(2019)Chalkidis, Androutsopoulos, and Aletras]{Chalkidis2019ACL}
Ilias Chalkidis, Ion Androutsopoulos, and Nikolaos Aletras.
\newblock Neural legal judgment prediction in {E}nglish.
\newblock In \emph{Annual Meeting of the Association for Computational Linguistics (ACL)}, 2019.

\bibitem[Chalkidis et~al.(2021)Chalkidis, Fergadiotis, Tsarapatsanis, Aletras, Androutsopoulos, and Malakasiotis]{Chalkidis2021ACL}
Ilias Chalkidis, Manos Fergadiotis, Dimitrios Tsarapatsanis, Nikolaos Aletras, Ion Androutsopoulos, and Prodromos Malakasiotis.
\newblock Paragraph-level rationale extraction through regularization: A case study on {E}uropean court of human rights cases.
\newblock In \emph{Annual Meeting of the Association for Computational Linguistics (ACL)}, pp.\  226--241, 2021.

\bibitem[Chalkidis et~al.(2022)Chalkidis, Dai, Fergadiotis, Malakasiotis, and Elliott]{Chalkidis2022ARXIV}
Ilias Chalkidis, Xiang Dai, Manos Fergadiotis, Prodromos Malakasiotis, and Desmond Elliott.
\newblock An exploration of hierarchical attention transformers for efficient long document classification.
\newblock \emph{arXiv.org}, 2210.05529, 2022.

\bibitem[Chen et~al.(2022)Chen, Chu, Wiseman, and Gimpel]{Chen2022ACL}
Mingda Chen, Zewei Chu, Sam Wiseman, and Kevin Gimpel.
\newblock Summscreen: {A} dataset for abstractive screenplay summarization.
\newblock In \emph{Annual Meeting of the Association for Computational Linguistics (ACL)}, 2022.

\bibitem[Dai et~al.(2022)Dai, Chalkidis, Darkner, and Elliott]{Dai2022EMNLP}
Xiang Dai, Ilias Chalkidis, Sune Darkner, and Desmond Elliott.
\newblock Revisiting transformer-based models for long document classification.
\newblock In \emph{Conference on Empirical Methods in Natural Language Processing (EMNLP)}, 2022.

\bibitem[Dao(2023)]{Dao2023ARXIV}
Tri Dao.
\newblock Flashattention-2: Faster attention with better parallelism and work partitioning.
\newblock \emph{arXiv.org}, 2307.08691, 2023.

\bibitem[Dao et~al.(2022)Dao, Fu, Ermon, Rudra, and R{\'{e}}]{Dao2022NEURIPS}
Tri Dao, Daniel~Y. Fu, Stefano Ermon, Atri Rudra, and Christopher R{\'{e}}.
\newblock Flashattention: Fast and memory-efficient exact attention with io-awareness.
\newblock In \emph{Advances in Neural Information Processing Systems (NeurIPS)}, 2022.

\bibitem[Dasigi et~al.(2021)Dasigi, Lo, Beltagy, Cohan, Smith, and Gardner]{Dasigi2021NAACL}
Pradeep Dasigi, Kyle Lo, Iz~Beltagy, Arman Cohan, Noah~A. Smith, and Matt Gardner.
\newblock A dataset of information-seeking questions and answers anchored in research papers.
\newblock In \emph{North American Chapter of the Association for Computational Linguistics (NAACL)}, 2021.

\bibitem[Devlin et~al.(2019)Devlin, Chang, Lee, and Toutanova]{Devlin2019NAACL}
Jacob Devlin, Ming-Wei Chang, Kenton Lee, and Kristina Toutanova.
\newblock {BERT:} pre-training of deep bidirectional transformers for language understanding.
\newblock In \emph{North American Chapter of the Association for Computational Linguistics (NAACL)}, 2019.

\bibitem[Ferris \& Horn(1998)Ferris and Horn]{Ferris1998MP}
Michael~C. Ferris and Jeffrey~D. Horn.
\newblock Partitioning mathematical programs for parallel solution.
\newblock \emph{Mathematical Programming (MP)}, 80:\penalty0 35--61, 1998.

\bibitem[Geiping \& Goldstein(2023)Geiping and Goldstein]{Geiping2023ICML}
Jonas Geiping and Tom Goldstein.
\newblock Cramming: Training a language model on a single {GPU} in one day.
\newblock In \emph{Proc. of the International Conf. on Machine learning (ICML)}, 2023.

\bibitem[Guthrie et~al.(1991)Guthrie, Britten, and Barker]{Guthrie1991RRQ}
John~T. Guthrie, Tracy Britten, and K.~Georgene Barker.
\newblock Roles of {Document} {Structure}, {Cognitive} {Strategy}, and {Awareness} in {Searching} for {Information}.
\newblock \emph{Reading Research Quarterly}, 26\penalty0 (3), 1991.

\bibitem[Habernal et~al.(2023)Habernal, Faber, Recchia, Bretthauer, Gurevych, Spiecker~genannt D{\"o}hmann, and Burchard]{Habernal2023AIandLaw}
Ivan Habernal, Daniel Faber, Nicola Recchia, Sebastian Bretthauer, Iryna Gurevych, Indra Spiecker~genannt D{\"o}hmann, and Christoph Burchard.
\newblock Mining legal arguments in court decisions.
\newblock \emph{Artificial Intelligence and Law}, 2023.

\bibitem[He et~al.(2016)He, Zhang, Ren, and Sun]{He2016CVPR}
Kaiming He, Xiangyu Zhang, Shaoqing Ren, and Jian Sun.
\newblock Deep residual learning for image recognition.
\newblock In \emph{Proc. IEEE Conf. on Computer Vision and Pattern Recognition (CVPR)}, 2016.

\bibitem[Hendrickson \& Kolda(2000)Hendrickson and Kolda]{Hendrickson2000SICS}
Bruce Hendrickson and Tamara~G. Kolda.
\newblock Partitioning rectangular and structurally unsymmetric sparse matrices for parallel processing.
\newblock \emph{SIAM Journal on Scientific Computing (SICS)}, 21\penalty0 (6):\penalty0 2048--2072, 2000.

\bibitem[Hong et~al.(2022)Hong, Kim, Kang, and Myaeng]{Hong2022EMNLP}
Giwon Hong, Jeonghwan Kim, Junmo Kang, and Sung{-}Hyon Myaeng.
\newblock Graph-induced transformers for efficient multi-hop question answering.
\newblock In \emph{Conference on Empirical Methods in Natural Language Processing (EMNLP)}, 2022.

\bibitem[Huang et~al.(2021)Huang, Cao, Parulian, Ji, and Wang]{Huang2021NAACL}
Luyang Huang, Shuyang Cao, Nikolaus~Nova Parulian, Heng Ji, and Lu~Wang.
\newblock Efficient attentions for long document summarization.
\newblock In \emph{North American Chapter of the Association for Computational Linguistics (NAACL)}, 2021.

\bibitem[Johnson et~al.(2016)Johnson, Pollard, Shen, Lehman, Feng, Ghassemi, Moody, Szolovits, Anthony~Celi, and Mark]{Johnson2016Nature}
Alistair E.~W. Johnson, Tom~J. Pollard, Lu~Shen, Li-wei~H. Lehman, Mengling Feng, Mohammad Ghassemi, Benjamin Moody, Peter Szolovits, Leo Anthony~Celi, and Roger~G. Mark.
\newblock Mimic-iii, a freely accessible critical care database.
\newblock \emph{Nature}, 3, 2016.

\bibitem[Kitaev et~al.(2020)Kitaev, Kaiser, and Levskaya]{Kitaev2020ICLR}
Nikita Kitaev, Lukasz Kaiser, and Anselm Levskaya.
\newblock Reformer: The efficient transformer.
\newblock In \emph{Proc. of the International Conf. on Learning Representations (ICLR)}, 2020.

\bibitem[Kocisk{\'{y}} et~al.(2018)Kocisk{\'{y}}, Schwarz, Blunsom, Dyer, Hermann, Melis, and Grefenstette]{Kocisky2018TACL}
Tom{\'{a}}s Kocisk{\'{y}}, Jonathan Schwarz, Phil Blunsom, Chris Dyer, Karl~Moritz Hermann, G{\'{a}}bor Melis, and Edward Grefenstette.
\newblock The narrativeqa reading comprehension challenge.
\newblock \emph{Transactions of the Association for Computational Linguistics (TACL)}, 6:\penalty0 317--328, 2018.

\bibitem[Koreeda \& Manning(2021)Koreeda and Manning]{Koreeda2021EMNLP}
Yuta Koreeda and Christopher~D. Manning.
\newblock Contractnli: {A} dataset for document-level natural language inference for contracts.
\newblock In \emph{Conference on Empirical Methods in Natural Language Processing (EMNLP)}, 2021.

\bibitem[Lewis et~al.(2020)Lewis, Liu, Goyal, Ghazvininejad, Mohamed, Levy, Stoyanov, and Zettlemoyer]{Lewis2022ACL}
Mike Lewis, Yinhan Liu, Naman Goyal, Marjan Ghazvininejad, Abdelrahman Mohamed, Omer Levy, Veselin Stoyanov, and Luke Zettlemoyer.
\newblock {BART:} denoising sequence-to-sequence pre-training for natural language generation, translation, and comprehension.
\newblock In \emph{Annual Meeting of the Association for Computational Linguistics (ACL)}, 2020.

\bibitem[Liu et~al.(2023)Liu, Lin, Hewitt, Paranjape, Bevilacqua, Petroni, and Liang]{Liu2023ARXIV}
Nelson~F. Liu, Kevin Lin, John Hewitt, Ashwin Paranjape, Michele Bevilacqua, Fabio Petroni, and Percy Liang.
\newblock Lost in the middle: How language models use long contexts.
\newblock \emph{arXiv.org}, 2307.03172, 2023.

\bibitem[Liu et~al.(2022)Liu, Liu, Chen, Lu, Feng, Feng, Sun, Tian, Wu, and Wang]{Liu2022ARXIVb}
Yang Liu, Jiaxiang Liu, Li~Chen, Yuxiang Lu, Shikun Feng, Zhida Feng, Yu~Sun, Hao Tian, Hua Wu, and Haifeng Wang.
\newblock {ERNIE-SPARSE:} learning hierarchical efficient transformer through regularized self-attention.
\newblock \emph{ARXIV}, abs/2203.12276, 2022.

\bibitem[Liu et~al.(2021)Liu, Zhang, Wan, Xia, He, and Yu]{Liu2021EMNLP}
Ye~Liu, Jian{-}Guo Zhang, Yao Wan, Congying Xia, Lifang He, and Philip~S. Yu.
\newblock {HETFORMER:} heterogeneous transformer with sparse attention for long-text extractive summarization.
\newblock In \emph{Conference on Empirical Methods in Natural Language Processing (EMNLP)}, 2021.

\bibitem[Liu et~al.(2019)Liu, Ott, Goyal, Du, Joshi, Chen, Levy, Lewis, Zettlemoyer, and Stoyanov]{Liu2019ARXIV1}
Yinhan Liu, Myle Ott, Naman Goyal, Jingfei Du, Mandar Joshi, Danqi Chen, Omer Levy, Mike Lewis, Luke Zettlemoyer, and Veselin Stoyanov.
\newblock Roberta: {A} robustly optimized {BERT} pretraining approach.
\newblock \emph{ARXIV}, abs/1907.11692, 2019.

\bibitem[Loshchilov \& Hutter(2019)Loshchilov and Hutter]{Loshchilov2019ICLR}
Ilya Loshchilov and Frank Hutter.
\newblock Decoupled weight decay regularization.
\newblock In \emph{Proc. of the International Conf. on Learning Representations (ICLR)}, 2019.

\bibitem[Matteo et~al.(2023)Matteo, Daniele, Martin, and François]{Matteo2023ARXIV}
Pagliardini Matteo, Paliotta Daniele, Jaggi Martin, and Fleuret François.
\newblock Faster causal attention over large sequences through sparse flash attention.
\newblock \emph{arXiv.org}, 2306.01160, 2023.

\bibitem[Meng et~al.(2021)Meng, Thaker, Zhang, Dong, Yuan, Wang, and He]{Meng2021ACL}
Rui Meng, Khushboo Thaker, Lei Zhang, Yue Dong, Xingdi Yuan, Tong Wang, and Daqing He.
\newblock Bringing structure into summaries: a faceted summarization dataset for long scientific documents.
\newblock In \emph{Annual Meeting of the Association for Computational Linguistics (ACL)}, 2021.

\bibitem[Milakov \& Gimelshein(2018)Milakov and Gimelshein]{Milakov2018ARXIV}
Maxim Milakov and Natalia Gimelshein.
\newblock Online normalizer calculation for softmax.
\newblock \emph{arXiv.org}, 1805.02867, 2018.

\bibitem[Murty et~al.(2023)Murty, Sharma, Andreas, and Manning]{Murty2023ACL}
Shikhar Murty, Pratyusha Sharma, Jacob Andreas, and Christopher Manning.
\newblock Grokking of hierarchical structure in vanilla transformers.
\newblock In \emph{Annual Meeting of the Association for Computational Linguistics (ACL)}, 2023.

\bibitem[Nikita \& R.(2018)Nikita and R.]{Nikita2018NAACL}
Nangia Nikita and Bowman~Samuel R.
\newblock {ListOps:} a diagnostic dataset for latent tree learning.
\newblock In \emph{North American Chapter of the Association for Computational Linguistics (NAACL)}, 2018.

\bibitem[Ostendorff et~al.(2022)Ostendorff, Rethmeier, Augenstein, Gipp, and Rehm]{Ostendorff2022EMNLP}
Malte Ostendorff, Nils Rethmeier, Isabelle Augenstein, Bela Gipp, and Georg Rehm.
\newblock Neighborhood contrastive learning for scientific document representations with citation embeddings.
\newblock In \emph{Conference on Empirical Methods in Natural Language Processing (EMNLP)}, 2022.

\bibitem[Pang et~al.(2022)Pang, Parrish, Joshi, Nangia, Phang, Chen, Padmakumar, Ma, Thompson, He, and Bowman]{Pang2022NAACL}
Richard~Yuanzhe Pang, Alicia Parrish, Nitish Joshi, Nikita Nangia, Jason Phang, Angelica Chen, Vishakh Padmakumar, Johnny Ma, Jana Thompson, He~He, and Samuel~R. Bowman.
\newblock Quality: Question answering with long input texts, yes!
\newblock In \emph{North American Chapter of the Association for Computational Linguistics (NAACL)}, 2022.

\bibitem[Raffel et~al.(2020)Raffel, Shazeer, Roberts, Lee, Narang, Matena, Zhou, Li, and Liu]{Raffel2020JMLR}
Colin Raffel, Noam Shazeer, Adam Roberts, Katherine Lee, Sharan Narang, Michael Matena, Yanqi Zhou, Wei Li, and Peter~J Liu.
\newblock Exploring the limits of transfer learning with a unified text-to-text transformer.
\newblock \emph{Journal of Machine Learning Research (JMLR)}, 21\penalty0 (140):\penalty0 1--67, 2020.

\bibitem[{\v R}eh{\r u}{\v r}ek \& Sojka(2010){\v R}eh{\r u}{\v r}ek and Sojka]{Rehurek2010LRECWORK}
Radim {\v R}eh{\r u}{\v r}ek and Petr Sojka.
\newblock Software framework for topic modelling with large corpora.
\newblock In \emph{International Conference on Language Resources and Evaluation (LREC) Workshops}, 2010.

\bibitem[Rewon et~al.(2019)Rewon, Scott, Alec, and Ilya]{Rewon2019ARXIV}
Child Rewon, Gray Scott, Radford Alec, and Sutskever Ilya.
\newblock Generating long sequences with sparse transformers.
\newblock \emph{arXiv.org}, 2019.

\bibitem[Ruan et~al.(2022)Ruan, Ostendorff, and Rehm]{Ruan2022ACL}
Qian Ruan, Malte Ostendorff, and Georg Rehm.
\newblock Histruct+: Improving extractive text summarization with hierarchical structure information.
\newblock In \emph{Annual Meeting of the Association for Computational Linguistics (ACL)}, 2022.

\bibitem[Sachan et~al.(2021)Sachan, Zhang, Qi, and Hamilton]{Sachan2020EACL}
Devendra Sachan, Yuhao Zhang, Peng Qi, and William~L. Hamilton.
\newblock Do syntax trees help pre-trained transformers extract information?
\newblock In \emph{Conference of the European Chapter of the Association for Computational Linguistics (EACL)}, 2021.

\bibitem[Saier et~al.(2023)Saier, Krause, and F{\"{a}}rber]{Saier2023JCDL}
Tarek Saier, Johan Krause, and Michael F{\"{a}}rber.
\newblock unarxive 2022: All arxiv publications pre-processed for nlp, including structured full-text and citation network.
\newblock In \emph{ACM/IEEE Joint Conference on Digital Libraries, {JCDL} (JCDL)}, 2023.

\bibitem[Sennrich et~al.(2016)Sennrich, Haddow, and Birch]{Sennrich2016ACL}
Rico Sennrich, Barry Haddow, and Alexandra Birch.
\newblock Neural machine translation of rare words with subword units.
\newblock In \emph{Annual Meeting of the Association for Computational Linguistics (ACL)}, 2016.

\bibitem[Shaham et~al.(2022)Shaham, Segal, Ivgi, Efrat, Yoran, Haviv, Gupta, Xiong, Geva, Berant, and Levy]{Shaham2022EMNLP}
Uri Shaham, Elad Segal, Maor Ivgi, Avia Efrat, Ori Yoran, Adi Haviv, Ankit Gupta, Wenhan Xiong, Mor Geva, Jonathan Berant, and Omer Levy.
\newblock {SCROLLS:} standardized comparison over long language sequences.
\newblock In \emph{Conference on Empirical Methods in Natural Language Processing (EMNLP)}, 2022.

\bibitem[Singh et~al.(2023)Singh, D'Arcy, Cohan, Downey, and Feldman]{Singh2023EMNLP}
Amanpreet Singh, Mike D'Arcy, Arman Cohan, Doug Downey, and Sergey Feldman.
\newblock Scirepeval: {A} multi-format benchmark for scientific document representations.
\newblock In \emph{Conference on Empirical Methods in Natural Language Processing (EMNLP)}, 2023.

\bibitem[Suzgun et~al.(2023)Suzgun, Melas-Kyriazi, Sarkar, Kominers, and Shieber]{Suzgun2023NEURIPSDATA}
Mirac Suzgun, Luke Melas-Kyriazi, Suproteem~K Sarkar, Scott Kominers, and Stuart Shieber.
\newblock The harvard {USPTO} patent dataset: A large-scale, well-structured, and multi-purpose corpus of patent applications.
\newblock In \emph{Proc. of the Neural Information Processing Systems (NeurIPS) Track on Datasets and Benchmarks}, 2023.

\bibitem[Tay et~al.(2021)Tay, Dehghani, Abnar, Shen, Bahri, Pham, Rao, Yang, Ruder, and Metzler]{Tay2021ICLR}
Yi~Tay, Mostafa Dehghani, Samira Abnar, Yikang Shen, Dara Bahri, Philip Pham, Jinfeng Rao, Liu Yang, Sebastian Ruder, and Donald Metzler.
\newblock Long range arena : {A} benchmark for efficient transformers.
\newblock In \emph{Proc. of the International Conf. on Learning Representations (ICLR)}, 2021.

\bibitem[Tay et~al.(2023{\natexlab{a}})Tay, Dehghani, Bahri, and Metzler]{Tay2023CS}
Yi~Tay, Mostafa Dehghani, Dara Bahri, and Donald Metzler.
\newblock Efficient transformers: {A} survey.
\newblock \emph{{ACM} Computing Surveys}, 55\penalty0 (6):\penalty0 109:1--109:28, 2023{\natexlab{a}}.

\bibitem[Tay et~al.(2023{\natexlab{b}})Tay, Dehghani, Tran, Garcia, Wei, Wang, Chung, Bahri, Schuster, Zheng, Zhou, Houlsby, and Metzler]{Tay2023ICLR}
Yi~Tay, Mostafa Dehghani, Vinh~Q. Tran, Xavier Garcia, Jason Wei, Xuezhi Wang, Hyung~Won Chung, Dara Bahri, Tal Schuster, Huaixiu~Steven Zheng, Denny Zhou, Neil Houlsby, and Donald Metzler.
\newblock {UL2:} unifying language learning paradigms.
\newblock In \emph{Proc. of the International Conf. on Learning Representations (ICLR)}, 2023{\natexlab{b}}.

\bibitem[Taylor \& Beach(1984)Taylor and Beach]{Taylor1984RRQ}
Barbara~M. Taylor and Richard~W. Beach.
\newblock The {Effects} of {Text} {Structure} {Instruction} on {Middle}-{Grade} {Students}' {Comprehension} and {Production} of {Expository} {Text}.
\newblock \emph{Reading Research Quarterly}, 19\penalty0 (2):\penalty0 134--146, 1984.

\bibitem[Tillet et~al.(2019)Tillet, Kung, and Cox]{Tillet2019MAPL}
Philippe Tillet, Hsiang{-}Tsung Kung, and David~D. Cox.
\newblock Triton: an intermediate language and compiler for tiled neural network computations.
\newblock In \emph{ACM SIGPLAN International Workshop on Machine Learning and Programming Languages (MAPL)}, 2019.

\bibitem[Vaswani et~al.(2017)Vaswani, Shazeer, Parmar, Uszkoreit, Jones, Gomez, Kaiser, and Polosukhin]{Vaswani2017NIPS}
Ashish Vaswani, Noam Shazeer, Niki Parmar, Jakob Uszkoreit, Llion Jones, Aidan~N. Gomez, Lukasz Kaiser, and Illia Polosukhin.
\newblock Attention is all you need.
\newblock In \emph{Advances in Neural Information Processing Systems (NIPS)}, pp.\  5998--6008, 2017.

\bibitem[Wang et~al.(2019)Wang, Lee, and Chen]{Wang2019EMNLP}
Yaushian Wang, Hung-Yi Lee, and Yun-Nung Chen.
\newblock Tree transformer: Integrating tree structures into self-attention.
\newblock In \emph{Conference on Empirical Methods in Natural Language Processing (EMNLP)}, 2019.

\bibitem[Wu et~al.(2021{\natexlab{a}})Wu, Wu, Qi, and Huang]{Wu2021IJCNLP}
Chuhan Wu, Fangzhao Wu, Tao Qi, and Yongfeng Huang.
\newblock Hi-transformer: Hierarchical interactive transformer for efficient and effective long document modeling.
\newblock In \emph{International Joint Conference on Natural Language Processing (IJCNLP)}, 2021{\natexlab{a}}.

\bibitem[Wu et~al.(2021{\natexlab{b}})Wu, Zhao, and Zhang]{Zhang2021ACL}
Hongqiu Wu, Hai Zhao, and Min Zhang.
\newblock Code summarization with structure-induced transformer.
\newblock In \emph{Annual Meeting of the Association for Computational Linguistics (ACL)}, 2021{\natexlab{b}}.

\bibitem[Yang(2019)]{Yang2019ARXIV}
Liu Yang.
\newblock Fine-tune {BERT} for extractive summarization.
\newblock \emph{arXiv.org}, a10a6495723ea8c928680ecdd61714f5750586c3, 2019.

\bibitem[Yang et~al.(2016)Yang, Yang, Dyer, He, Smola, and Hovy]{Yang2016NAACL}
Zichao Yang, Diyi Yang, Chris Dyer, Xiaodong He, Alex Smola, and Eduard Hovy.
\newblock Hierarchical attention networks for document classification.
\newblock In \emph{North American Chapter of the Association for Computational Linguistics (NAACL)}, 2016.

\bibitem[Zaheer et~al.(2020)Zaheer, Guruganesh, Dubey, Ainslie, Alberti, Ontanon, Pham, Ravula, Wang, Yang, and Ahmed]{Zaheer2020NEURIPS}
Manzil Zaheer, Guru Guruganesh, Kumar~Avinava Dubey, Joshua Ainslie, Chris Alberti, Santiago Ontanon, Philip Pham, Anirudh Ravula, Qifan Wang, Li~Yang, and Amr Ahmed.
\newblock Big bird: Transformers for longer sequences.
\newblock In \emph{Advances in Neural Information Processing Systems (NeurIPS)}, 2020.

\bibitem[Zhang et~al.(2022)Zhang, Liu, and Zhang]{Zhang2022EMNLP}
Haopeng Zhang, Xiao Liu, and Jiawei Zhang.
\newblock {HEGEL:} hypergraph transformer for long document summarization.
\newblock In \emph{Conference on Empirical Methods in Natural Language Processing (EMNLP)}, 2022.

\bibitem[Zhong et~al.(2021)Zhong, Yin, Yu, Zaidi, Mutuma, Jha, Awadallah, Celikyilmaz, Liu, Qiu, and Radev]{Zhong2021NAACL}
Ming Zhong, Da~Yin, Tao Yu, Ahmad Zaidi, Mutethia Mutuma, Rahul Jha, Ahmed~Hassan Awadallah, Asli Celikyilmaz, Yang Liu, Xipeng Qiu, and Dragomir~R. Radev.
\newblock Qmsum: {A} new benchmark for query-based multi-domain meeting summarization.
\newblock In \emph{North American Chapter of the Association for Computational Linguistics (NAACL)}, 2021.

\end{thebibliography}
\bibliographystyle{colm_files/colm2024_conference}

\appendix

\section{Appendix}
\subsection{Hierarchical Attention Kernel}

\begin{figure}[h]
  \subfloat[Longformer's fixed-window attention]{
	\begin{minipage}[c][0.8\width]{
	   0.49\textwidth}
	   \centering
	   \includegraphics[width=1\textwidth]{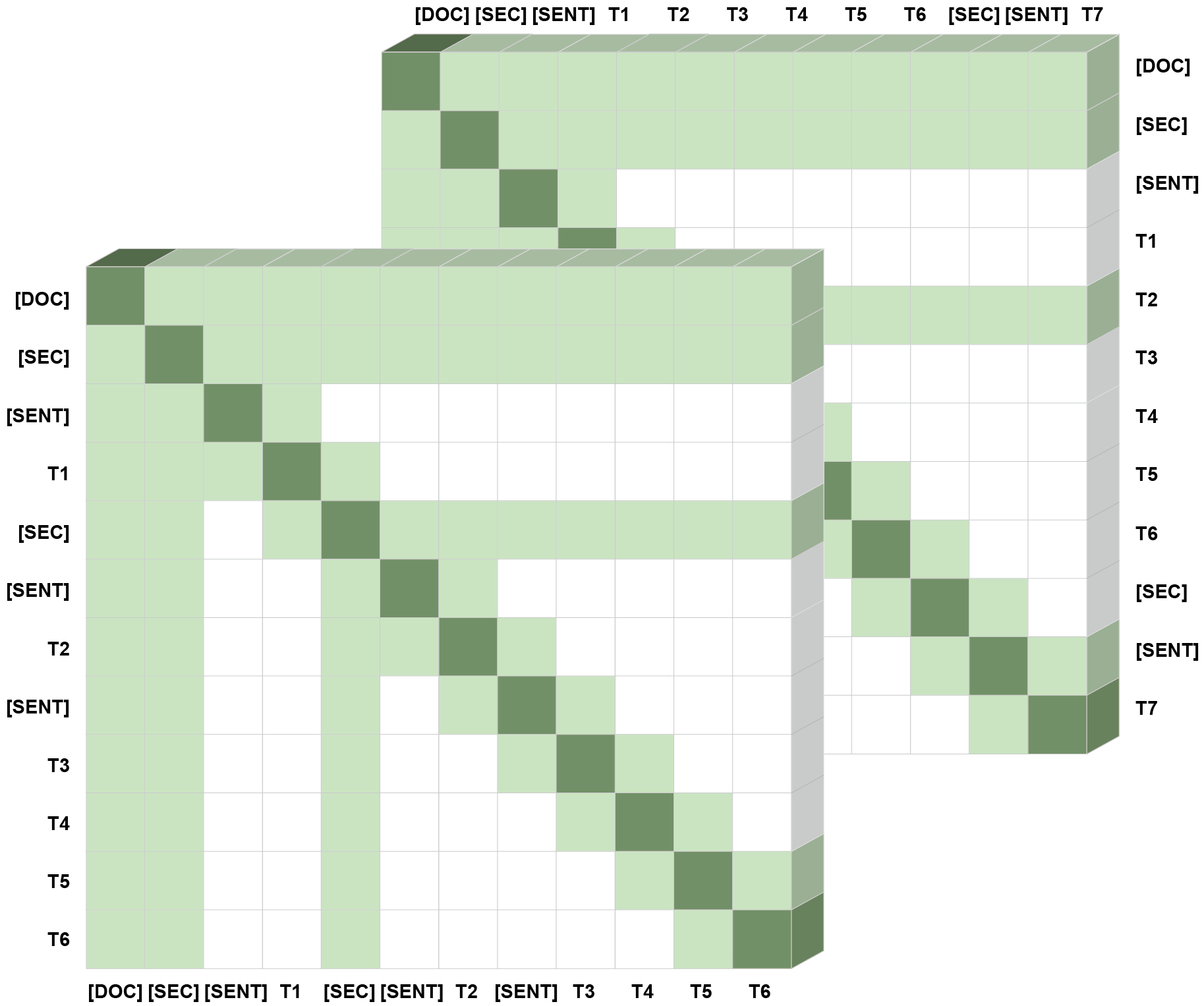}
	\end{minipage}}
 \hfill 	
  \subfloat[Our dynamic hierarchical attention]{
	\begin{minipage}[c][0.8\width]{
	   0.49\textwidth}
	   \centering
	   \includegraphics[width=1\textwidth]{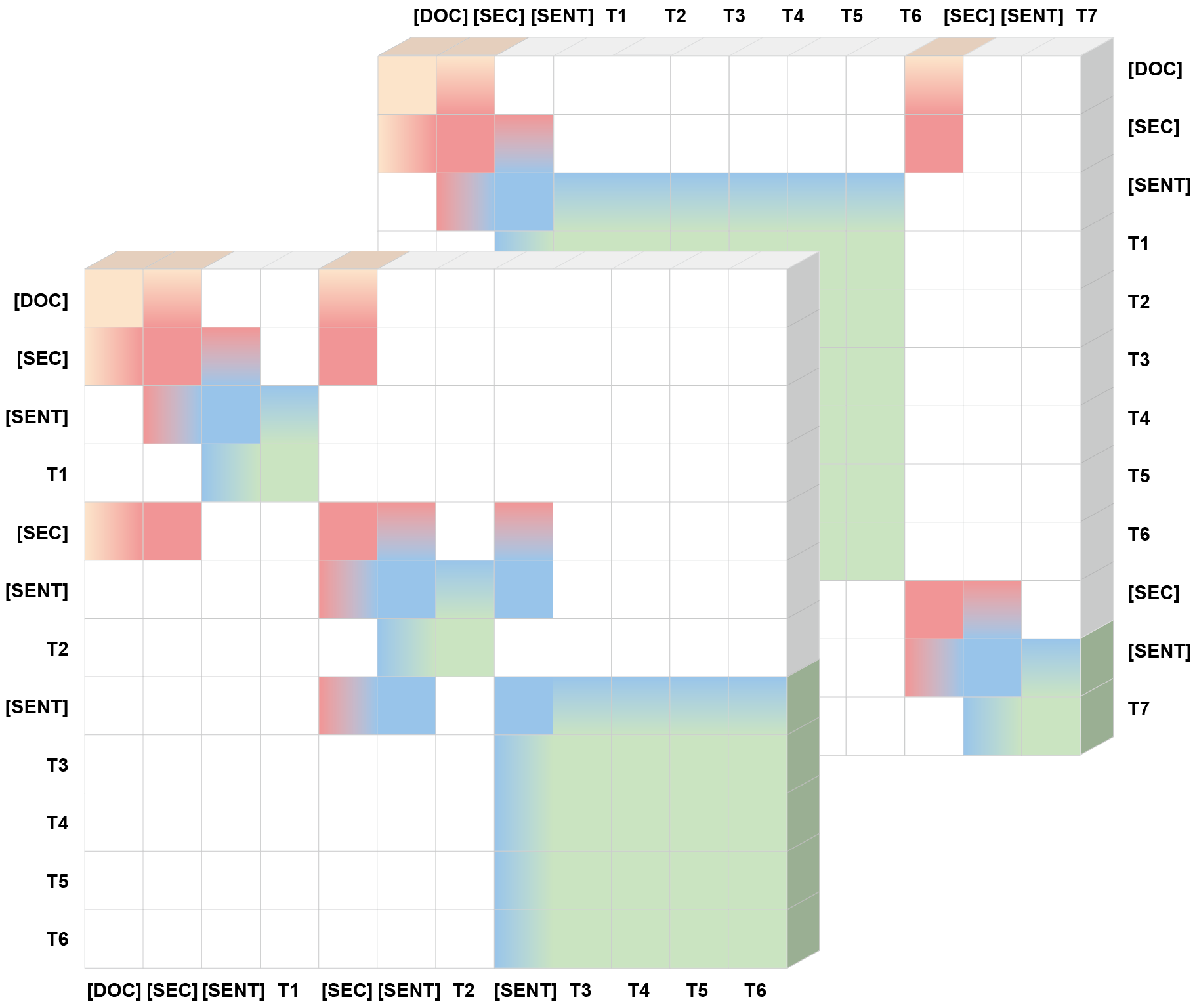}
        \label{fig: hierarchical_attn}
	\end{minipage}}
\caption{(a) The Longformer~\citep{Beltagy2020ARXIV} sparse attention pattern is identical for all samples in a mini-batch. (b) In contrast, the proposed dynamic hierarchical attention pattern considers the document structure and hence is different for each sample in a mini-batch.}
\label{fig:attn_diff}
\end{figure}

In this section, we compare the proposed dynamic hierarchical attention pattern to the fixed sparse attention pattern of Longformer and provide algorithmic details of its implementation. Subsequently, we illustrate how our customized attention kernel reduces computational overhead through qualitative examples.

\begin{figure}[h]
\vspace{0.5cm}
\begin{subfigure}[t]{0.48\textwidth}
    \includegraphics[width=\textwidth]{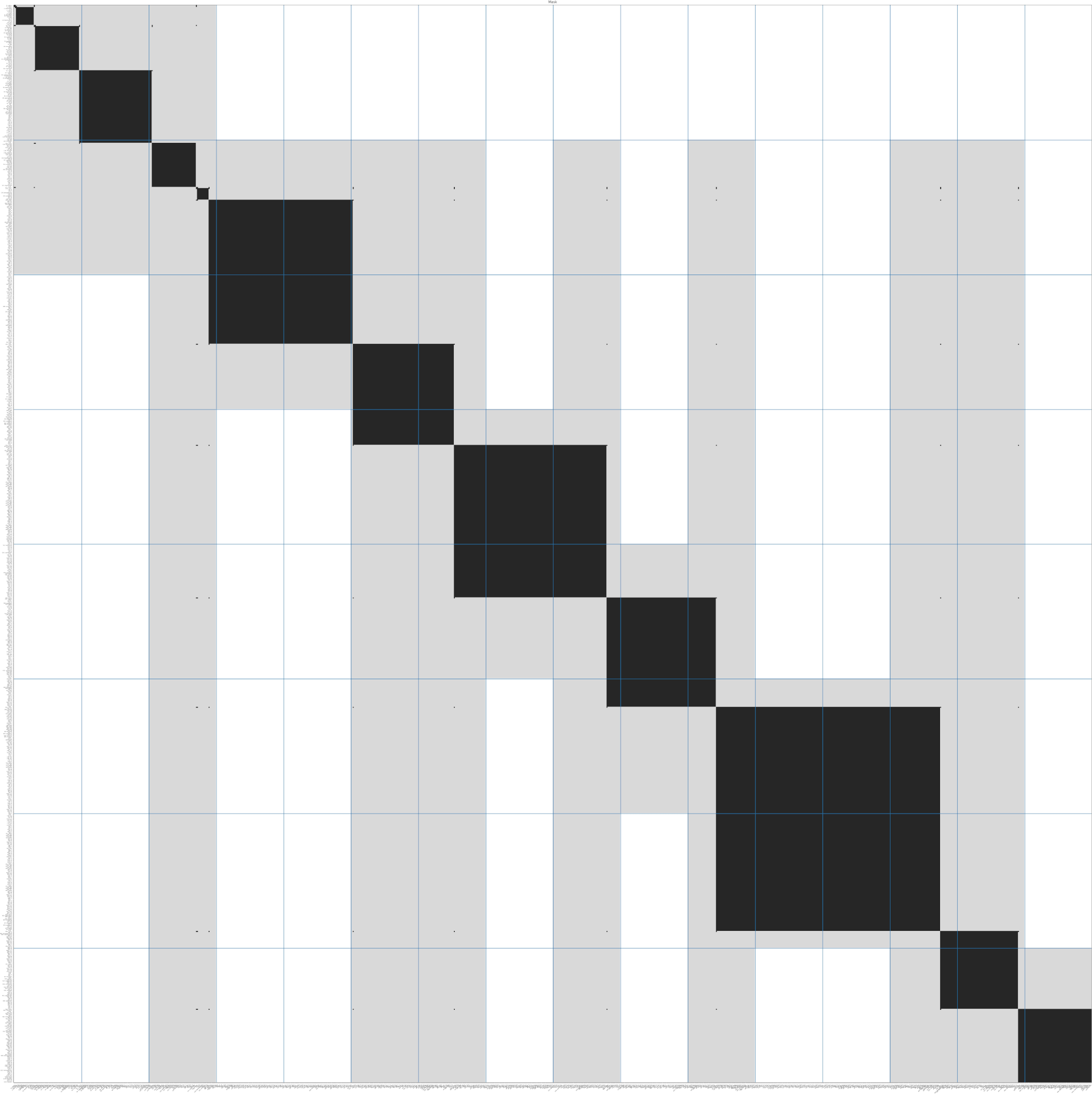}
    \caption{Block-Sparse FlashAttention~\citep{Dao2022NEURIPS}}
    \vspace{0.3cm}
\end{subfigure}\hspace{\fill} 
\begin{subfigure}[t]{0.48\textwidth}
    \includegraphics[width=\linewidth]{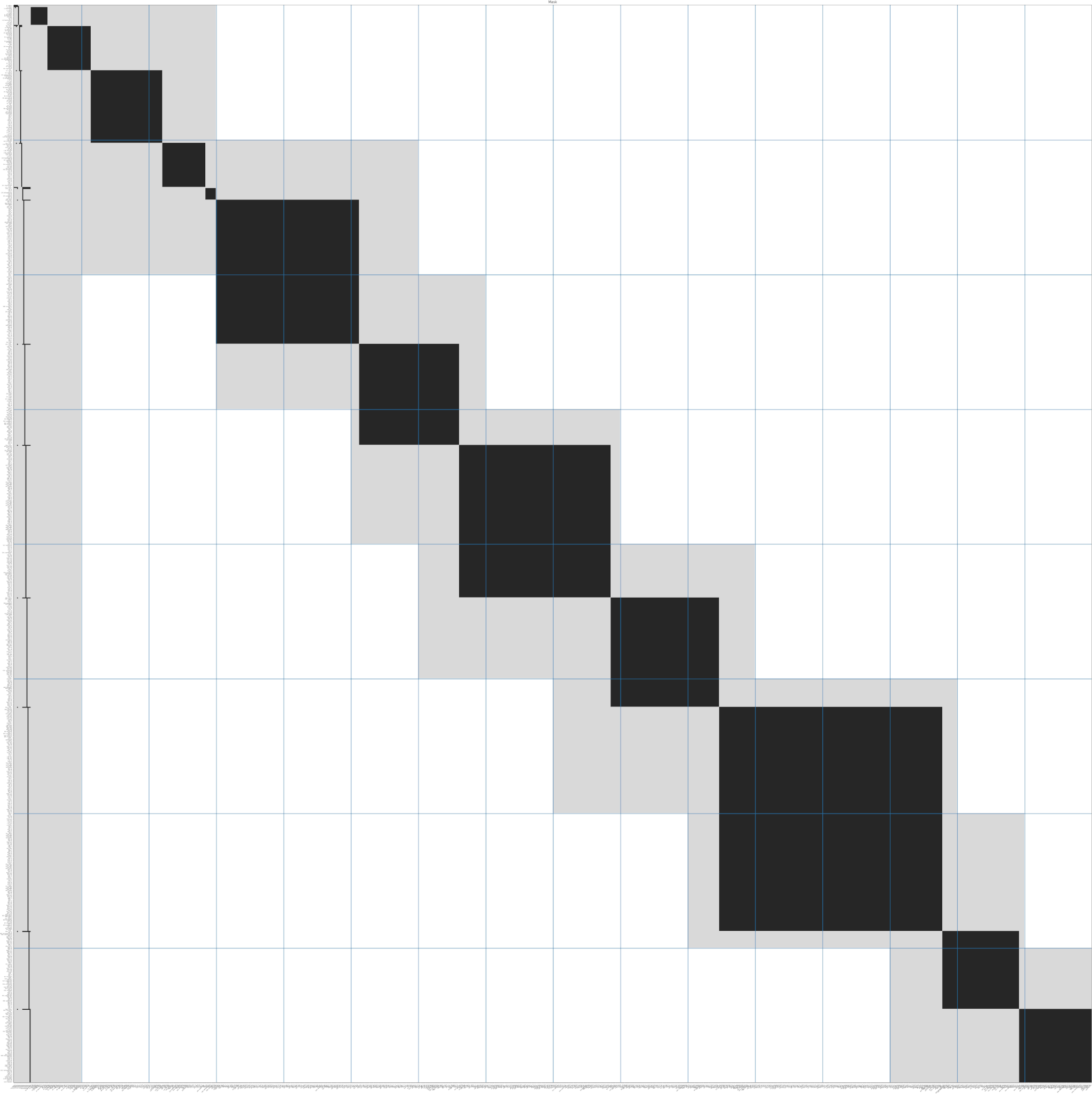}
    \caption{HDT Attention}
    \vspace{0.3cm}
\end{subfigure}
\begin{subfigure}[t]{0.48\textwidth}
    \includegraphics[width=\linewidth]{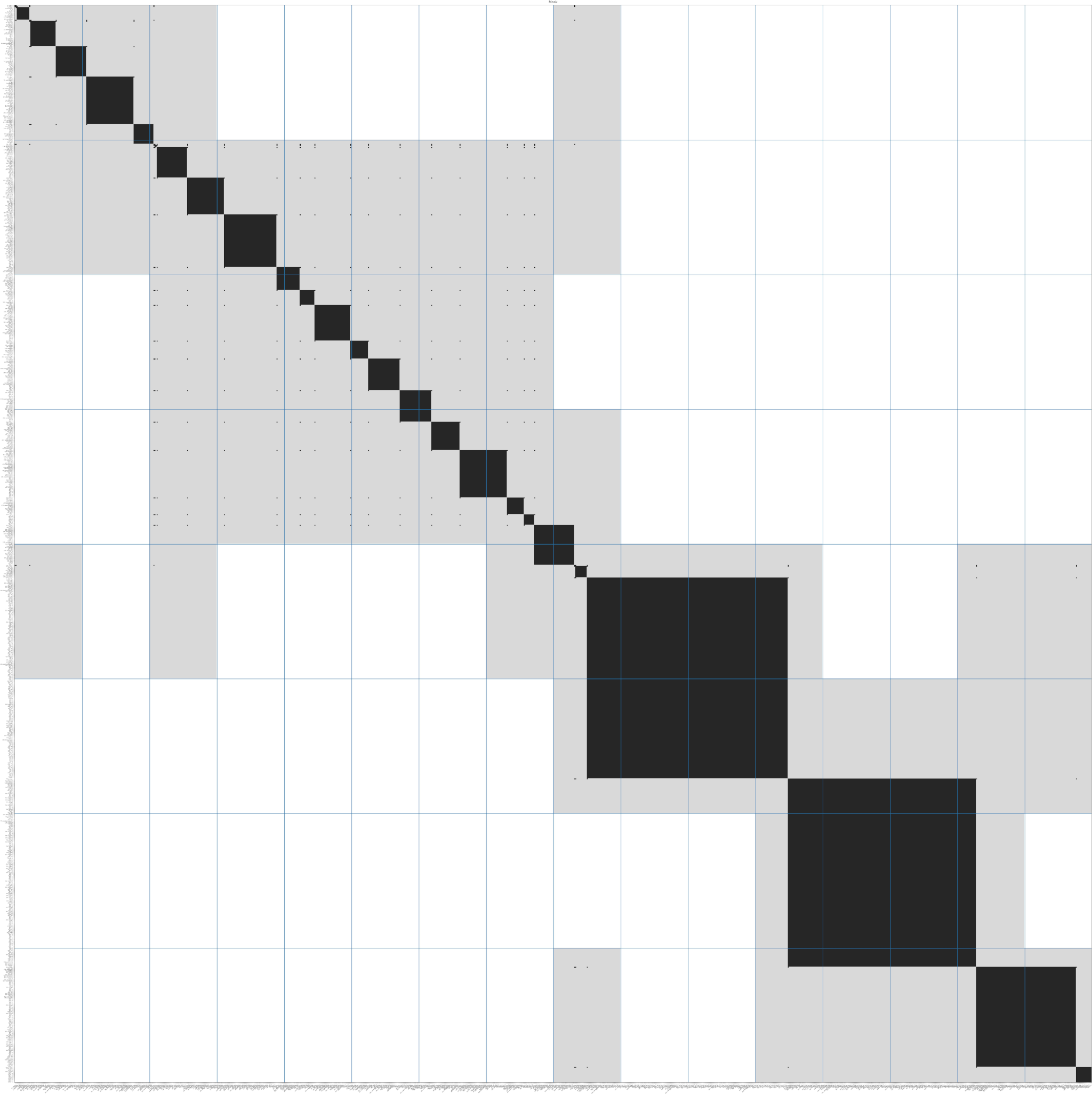}
    \caption{Block-Sparse FlashAttention~\citep{Dao2022NEURIPS}}
\end{subfigure}\hspace{\fill} 
\begin{subfigure}[t]{0.48\textwidth}
    \includegraphics[width=\linewidth]{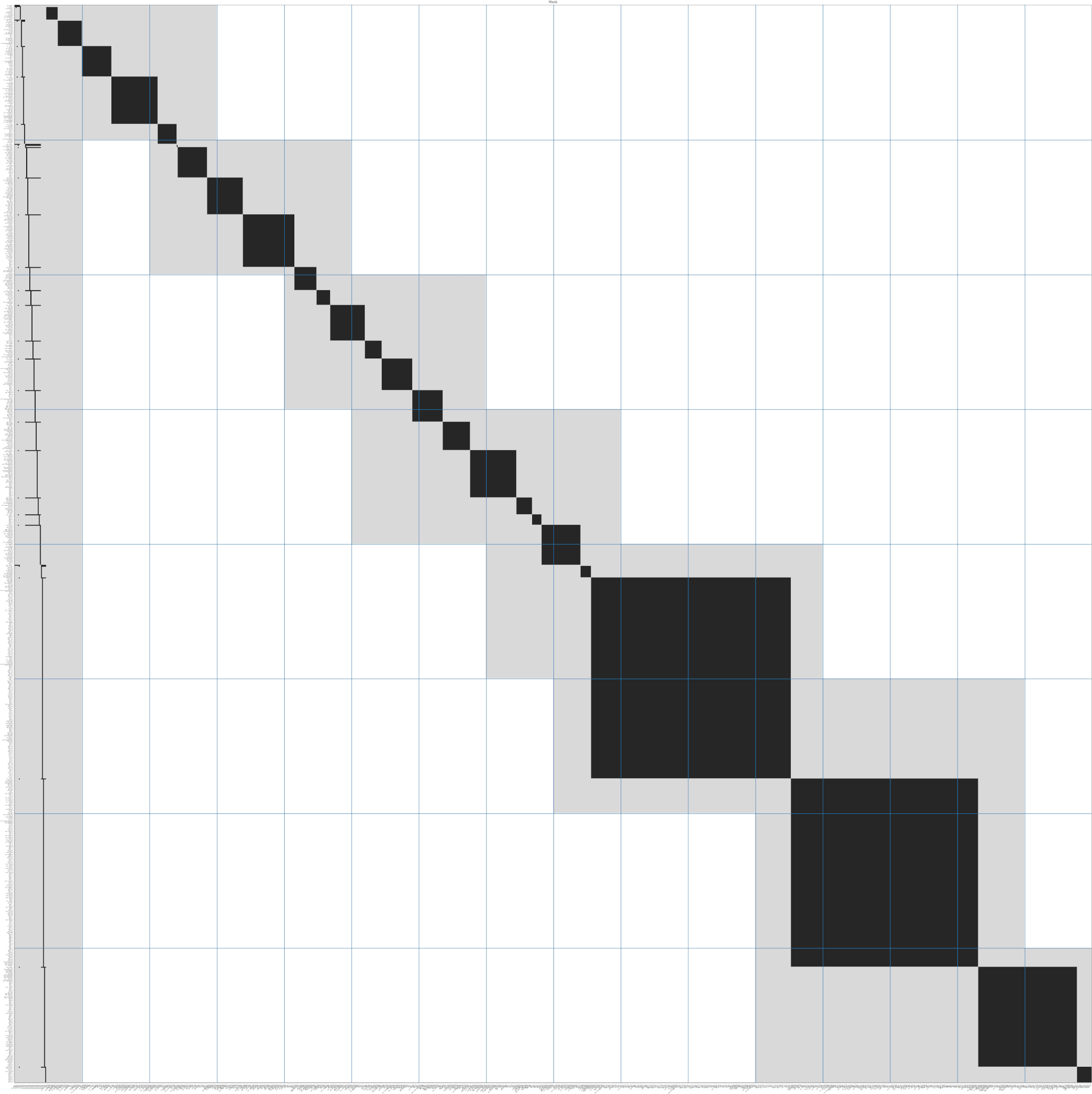}
    \caption{HDT Attention}
\end{subfigure}
\caption{\textbf{Attention Pattern (Black) and Processed SRAM Blocks (Grey).} Comparison between the practical computation of Block-Sparse FlashAttention~\citep{Dao2022NEURIPS} and our HDT attention kernel on the same hierarchical attention patterns. We show the attention mask of the first 1k tokens ($\sim$25\% of the total document size) of two different documents (row 1+2) in black.
The blue grid illustrates the $128 \times 64$ SRAM blocks which are processed in parallel using the fused kernel.
All blocks highlighted in grey contain at least one non-zero attention entry and hence require processing.
Due to the reordering of keys and values (columns) in HDT, anchor tokens are aggregated within adjacent blocks leading to a larger number of blocks that can be skipped compared to Block-Sparse FlashAttention~\citep{Dao2022NEURIPS}.}
\label{fig:full_attn_mask}
\end{figure}

Given that our attention pattern is rooted in document structure, it is different for each sample. Hence, we require an attention mechanism that is able to cope with sample-dependent attention patterns.
\figref{fig:attn_diff} provides a visual illustration comparing our dynamic attention pattern to the fixed attention pattern of Longformer~\citep{Beltagy2020ARXIV}, assuming a mini-batch size of 2 for clarity. Notably, within a mini-batch containing multiple samples, the attention pattern applied to each sample varies, thereby complicating the realization of our attention pattern using existing attention kernels. Moreover, due to the presence of anchor tokens positioned at the onset of different hierarchies and attending to one another, the attention devoted to these anchor tokens can be notably fine-grained and sparse within the attention mask. This phenomenon is demonstrated in \figref{fig:full_attn_mask}, depicting hierarchical attention masks derived from real documents. Note that due to space limitations only the first 1k tokens ($\sim$25\% of the total document size) are visualized. To optimize the number of empty blocks that can be skipped, we employ the sorting heuristic outlined in \secref{sec:hdt}. The core mechanism of our hierarchical sparse attention kernel is based on and extends FlashAttention~\citet{Dao2022NEURIPS}. Our algorithm for a forward pass of the attention layer is provided in Algorithm \ref{alg:sparse_kernel}. This algorithm is executed in parallel for all elements of the Cartesian product $B \times H$, where $B$ represents the batch size and $H$ denotes the number of heads, assuming input tensors of shape $(B \times H \times n \times d_k)$.

Our primary adjustment to the FlashAttention algorithm is in line 2 of Algorithm\ref{alg:sparse_kernel}, where we implement a sorting mechanism for the key and value based on their hierarchy levels, specifically prioritizing [DOC] tokens followed by [SEC] and [SENT] tokens before normal tokens. This modification is grounded in the observation that the computational inefficiency in block-wise hierarchical attention predominantly stems from attention interactions between anchor tokens situated distantly from each other, resulting in excessively sparse blocks. \figref{fig:full_attn_mask} offers a visual comparison between the computation patterns of our kernel and block-sparse FlashAttention, demonstrating the efficacy of our approach.

\begin{algorithm}[h]
\caption{Forward Pass of the \modelname~Hierarchical Attention Kernel}
\label{alg:sparse_kernel}
\begin{algorithmic}[1]
\Require Query, key, value $\bQ, \bK, \bV \in \mathbb{R}^{n \times d_k}$,
query block size $M$, key-value block size $N$, softmax statistics vectors $\mathbf{m} \in \mathbb{R}^{n \times n}$ and $\mathbf{l} \in \mathbb{R}^{n \times n}$, output tensor $\mathbf{O}^{n \times d_k}$.
\State Partition $\bQ$ into $T_r=\left\lceil \frac{n}{M} \right\rceil$ blocks $\bQ_1, \dots, \bQ_{T_r}$ 
\State Sort $\bK, \bV$ according to the hierarchy level from $l=1$ to $L$. 
\State Partition $\bK, \bV$ into $T_c=\left\lceil \frac{n}{N} \right\rceil$ blocks $\bK_1, \dots, \bK_{T_c}$, and $\bV_1, \dots, \bV_{T_c}$ 
\For{$1 \leq i \leq T_r$}
\State Load $\bQ_i, \bO_i, m_i, l_i$ from HBM to SRAM
\For{$1 \leq j \leq T_c$}
\State Load $\bK_j, \bV_j$ from HBM to SRAM
\State On chip, compute mask $\bM$ according to Equation \eqref{eq:attn_mask}
\If{Non-zero values in $\bM$}
\State $\bS_{ij}=\bQ_i \bK_j \odot \bM_{ij}$
\State $\tilde{m}_{ij}=\text{rowmax}(\bS_{ij})$
\State $\tilde{\bP}_{ij}=\exp(\bS_{ij}-\tilde{m}_{ij})$ (pointwise)
\State $\tilde{l}_{ij}=\text{rowsum}(\tilde{\bP}_{ij})$
\State $m_i^{\text{new}}=\max (m_i, \tilde {m}_{ij}) \in \nR^{M}, l^{\text{new}}_i=e^{m_i-m_i^{\text{new}}} l_i + e^{\tilde{m}_{ij}-m_i^{\text{new}}} \tilde{l}_{i,j} \in \nR^{M}$
\State Write $\bO_i \gets \text{diag}(l_i^{new})^{-1}(\text{diag}(l_i)e^{m_i-m_i^{new}} \bO_i + e^{\tilde{m}_{ij}-m_i^{new}} \tilde{\bP_{ij}} \bV_j)$ to HBM
\State Write $l_i \gets l_i^{new}$, $m_i \gets m_i^{new}$ to HBM
\EndIf
\EndFor

\EndFor
\end{algorithmic}
\label{alg:hierarchical_attention_kernel}
\end{algorithm}

\clearpage

\subsection{Experiments}

In this section, we provide additional details about the model settings and datasets.

\subsubsection{Mathematical Reasoning Tasks}

\begin{figure}[ht] 
  \begin{tabular}{cc} 
      \centering
      \includegraphics[width=0.475\textwidth]{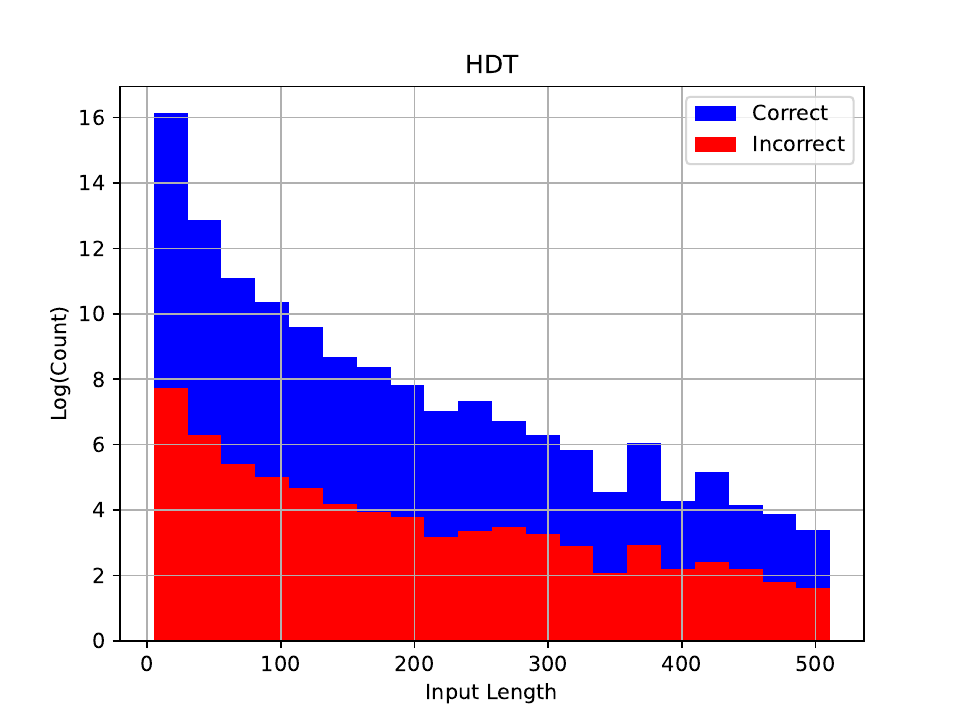}
    &
      \centering
      \includegraphics[width=0.475\textwidth]{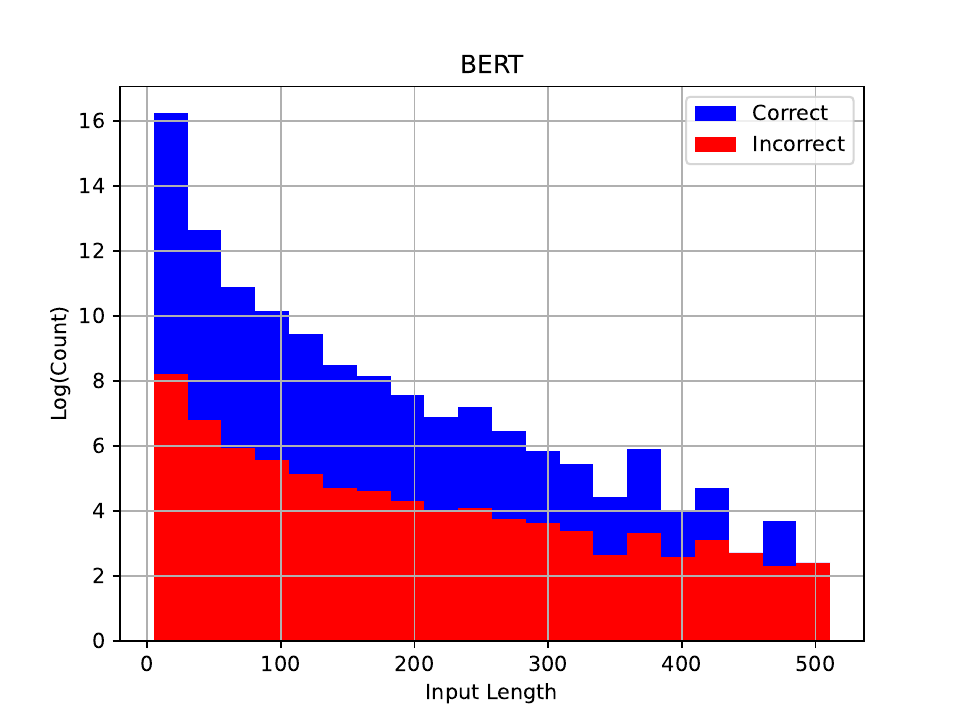}
    \cr
      \centering
      \includegraphics[width=0.475\textwidth]{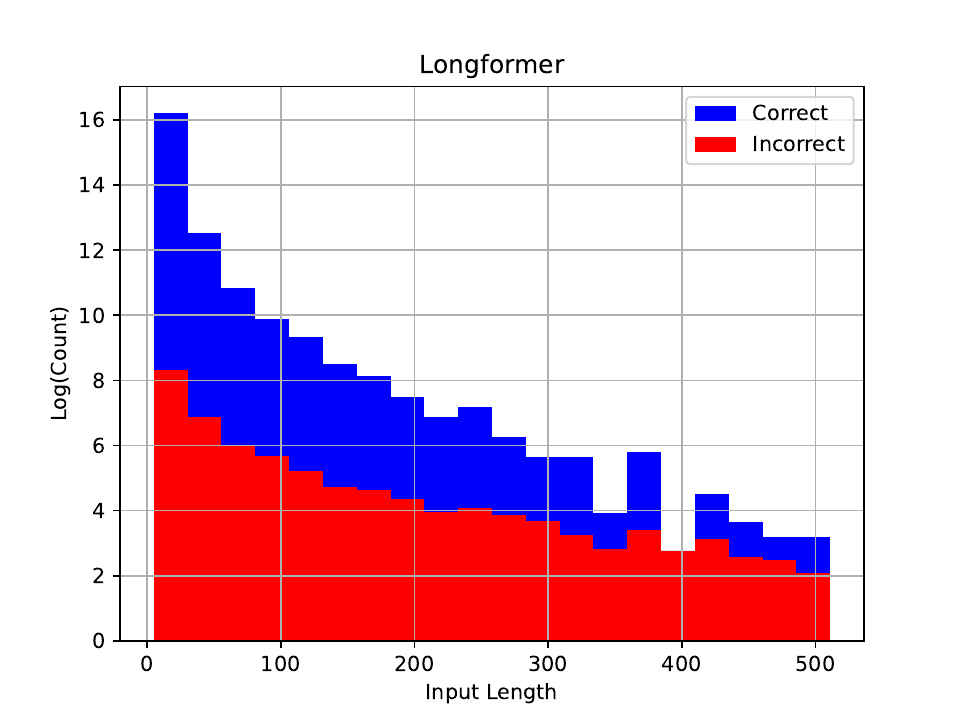}
    &
      \centering
      \includegraphics[width=0.475\textwidth]{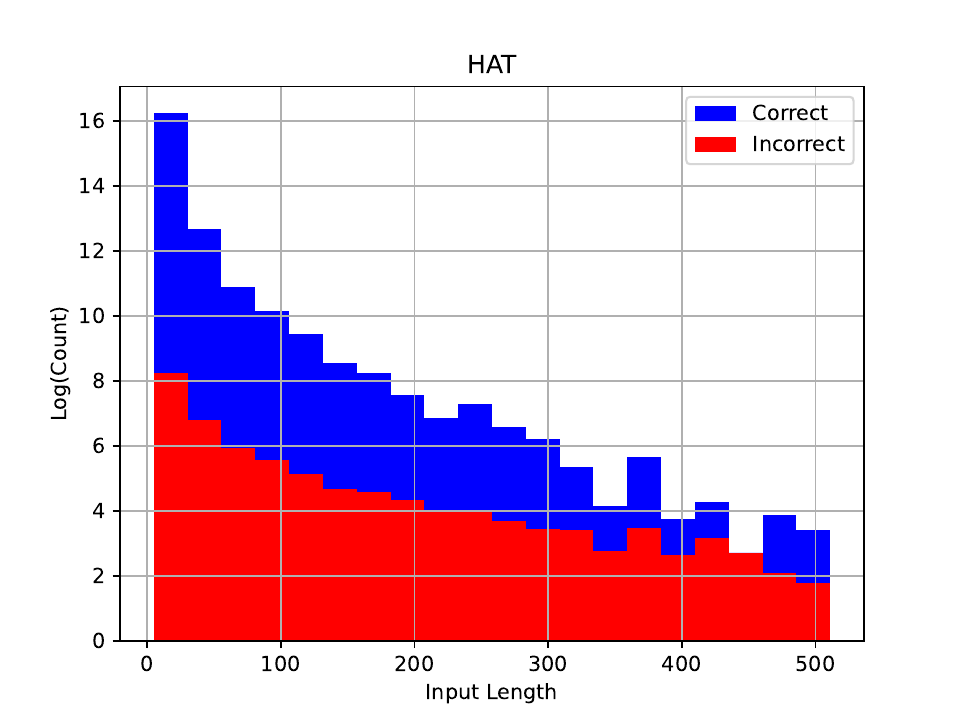}
  \end{tabular}

  \caption{Stacked histograms of predictions by HDT, BERT, Longformer and HAT on ListOPs 20d.}
  \label{fig:listops_histograms}

\end{figure}

\boldparagraph{ListOps}
The models we train on ListOPs 20d have 12 transformer encoder layers with a feature dimension of 128, an intermediate size of 512, a learning rate of 3e-4 and a batch size of 200. We split the original training set into 85k samples for training and 5k samples for validation. The test set accuracy is computed on 10k samples.
The ListOps code~\citep{Nikita2018NAACL} generates many short samples and a few very long ones, occasionally exceeding the 512 token input size of the BERT model. The largest token length that can be generated by ListOps 20d is $5^{20}$, with a maximum tree depth of 20 and the maximum number of operands per operator being 5. We exclude such rare samples from the dataset for our experiments.
We observe that our conclusions regarding the benefits of sparse structure-aware attention patterns also hold on ListOps data generated with a smaller maximum tree depth of 10d and 5d. 
As shown in \figref{fig:listops_histograms}, HDT is capable of correctly predicting long ListOPs sequences, whereas BERT, Longformer and HAT exhibit lower performance for long samples. HDT's performance on ListOPs is stable across varying input lengths.

\subsubsection{Language Tasks}

\boldparagraph{Model Settings}
\label{sec:settings}
We choose a typical setting for our model architecture, with $d_{\text{model}}=768$, $d_k=64$ with 12 heads, and an intermediate hidden size of 3,072. For \modelname-E, we use 12 encoder stacks, and for \modelname-ED we use 6 encoder stacks and 6 decoder stacks, resulting in 109M and 124M parameters, respectively. For fine-tuning, we use a constant learning rate of 5e-5, batch size 32 (by accumulating gradients of 8 mini-batches with mini-batch size 4), and a dropout rate of 0.1 for all tasks, except for summarization tasks for which we use a larger learning rate of 1e-3 which we empirically find better suited for all models in our experiments. We use AdamW \citep{Loshchilov2019ICLR} as the optimizer for both pre-training and fine-tuning. 

\boldparagraph{SciRepEval}
The SciRepEval benchmark we use in this work is a subset of the original dataset~\citep{Singh2023EMNLP} with additional full-text data from unarXive \citep{Saier2023JCDL}. We remove samples that are not in unarXive and drop the tasks that have less than 100 samples left after filtering. This leads to a benchmark spanning 13 tasks, including regression, proximity, and search tasks for scientific document representation evaluation. We show our results on proximity tasks in \tabref{tab:scirepeval} and results on the remaining tasks in \tabref{tab:reg}. 

\begin{table*}[t!]
\centering
\resizebox{1.0 \linewidth}{!}{%
\begin{tabular}{l|c|c|ccccc|c}
\toprule
Model & \multicolumn{1}{|c|}{SRH} & \multicolumn{1}{|c|}{PRX} & \multicolumn{5}{|c|}{RGN}  \\
       & Search & Feeds-1 & Peer Review Score &  Tweet Mentions  & hIndex & Citation Count & Publication Year & Average \\
            &   nDCG   & MAP  & Kendall’s $\mathcal{T}$ & Kendall’s $\mathcal{T}$  & Kendall’s $\mathcal{T}$      & Kendall’s $\mathcal{T}$ & Kendall’s $\mathcal{T}$ &         \\ \hline
$\mathtt{SciBERT}$ & 56.79 & 69.2   & \textbf{17.39} &	10.09 &	\textbf{16.84} &	15.43 &	13.37 &  28.44   \\
$\mathtt{Longformer}$ & 56.11 &	66.46   & 17.34 & 12.68 &	16.65 &	26.16 &	\textbf{25.43} &   31.54   \\
$\mathtt{SciNCL}$ & 56.05 &	\textbf{71.85}  & 13.15 &	\textbf{13.95} &	11.92 &	26.27 &	18.69 &  30.27   \\ 
$\mathtt{HAT}$ & \textbf{57.01} & 70.01  & 14.62 &	13.18 &	13.64 &	19.75 &	19.8 & 29.72    \\ \hline

$\mathtt{\modelname}$-$\mathtt{E}$ (title+abstract)  & 56.62 &	71.65 & 13.79	& 13.62 &	15.65 &	\textbf{26.34} &	24.66 &  \textbf{31.76}   \\

\bottomrule
\end{tabular}}
\caption{\textbf{Results on remaining SciRepEval tasks.} Although our model is not the best across all tasks, the performance is comparably stable in different tasks than other models.}
\label{tab:reg}
\end{table*}

\boldparagraph{Long Document Classification Tasks}
Following HAT \citep{Chalkidis2022ARXIV}, we validate our model on three additional long document classification tasks. The first task is MIMIC-III \citep{Johnson2016Nature}, which comprises nearly 50,000 discharge summaries from US hospitals. Each summary is annotated with one or more ICD-9 codes (labels). The model takes a discharge summary as input and outputs the relevant set of first-level ICD-9 codes (19 in total). The second task is ECtHR-LJP \citep{Chalkidis2021ACL}, containing approximately 11,000 cases from the European Court of Human Rights (ECtHR) public database. For each case, the dataset provides a list of factual paragraphs (facts) from the case description. Each case is mapped to articles of the ECHR that were allegedly violated. The model takes the list of facts as input and outputs the set of allegedly violated articles. Lastly, we include the ECtHR-ARG task \citep{Habernal2023AIandLaw}, which includes approximately 300 cases from the European Court of Human Rights (ECtHR). For each case, the dataset provides a list of argumentative paragraphs from the case analysis, with spans in each paragraph labeled with one or more of 13 argument types. We follow HAT in re-formulating this task as a sequential paragraph classification task, where each paragraph is labeled with one or more argument types. The model takes the list of paragraphs of a case as input and outputs the set of relevant argument types for each paragraph. Our results in \tabref{tab:hat_classification} show that our model is comparable to HAT on this task. 

\begin{table*}[t!]
\centering
\begin{tabular}{l|c|c|c}
\toprule
Model  & MIMIC & ECtHR-LJP & ECtHR-ARG \\
            &   F1   & F1  & acc  \\ \hline

$\mathtt{Longformer}$ & 78.7 &	 78.6   & 66.7  \\

$\mathtt{HAT}$ & 78.9 & \bf{79.8}  & 82.6   \\ \hline

$\mathtt{\modelname}$-$\mathtt{E}$  & \bf{79.1}  & 79.5 & \bf{82.9} \\

\bottomrule
\end{tabular}
\caption{\textbf{Results on document classification task ECtHR-LJP, ECtHR-ARG, MIMIC.} We follow the experimental setting of HAT and cover three long document classification tasks and report the test scores. }
\label{tab:hat_classification}
\end{table*}

\boldparagraph{FacetSum}
The FacetSum dataset provides fixed 4-class summaries, ``Purpose'', ``Method'', ``Findings'', and ``Value'', for each document. To align classes to the corresponding sections, we follow \citet{Meng2021ACL} to classify sections into ``Introduction'', ``Method'', ``Result'' and ``Conclusion'' first by keyword matching and then match the four classes of sections to the four classes of summaries, respectively. 

\boldparagraph{SCROLLS}
\label{sec:benchmarks}
The SCROLLS benchmark contains a suite of tasks that
require reasoning over long texts. The tasks cover GovReport \citep{Huang2021NAACL} and SummScreenFD \citep{Chen2022ACL} that are summarization tasks in the domain of government reports and TV shows; QMSum \citep{Zhong2021NAACL}, a query-based summarization task for meeting transcripts; QASPER \citep{Dasigi2021NAACL}, a question answering dataset for scientific papers; NarrativeQA \citep{Kocisky2018TACL}, a question answering dataset over entire books from Project Gutenberg\footnote{\url{https://www.gutenberg.org/}}; QuALITY \citep{Pang2022NAACL}, a multiple choice question answering dataset over stories
and articles sourced from Project Gutenberg; and Contract NLI \citep{Koreeda2021EMNLP} as a natural language inference dataset in the legal domain. \tabref{tab:scrolls_full} compares our model with SotA billion-parameter encoder-decoder models which have been trained on large industrial compute clusters.

\begin{table*}[t!]
\centering
\resizebox{1.0 \linewidth}{!}{%
\setlength{\tabcolsep}{3pt}
\begin{tabular}{l|ccccccc|c}
\toprule
\multirow{2}{*}{\textbf{Model}} &  \textbf{GovRep} & \textbf{SumScr} &  \textbf{QMSum}  & \textbf{Qspr} & \textbf{Nrtv} & \textbf{QALT} & \textbf{CNLI} & Avg \\
           &   ROUGE-1/2/L     & ROUGE-1/2/L & ROUGE-1/2/L  & F1      & F1 & EM-T/H  &  EM &       \\ \hline

$\mathtt{CoLT5}^\dag_{\mathtt{XL}}$ & 61.3/32.2/33.8   & 36.4/10.1/21.7
 & 36.2/12.9/24.2 &	53.9 &	31.1 &	48.1/43.8 & 88.4 &  43.51 \\
$\mathtt{LongT5}^\dag_{\mathtt{XL}}$ & 61.1/32.3/33.7   & 35.8/9.6/21.1
 & 34.9/11.8/23.5 &	53.1 &	29.3 &	46.0/42.1 & 88.2 &  42.53 \\
$\mathtt{CoLT5}^\dag_{\mathtt{Large}}$ & 60.7/31.3/32.9   & 36.7/10.6/22.0
 & 34.9/11.5/23.1 &	49.8 &	27.7 &	39.9/36.8 & 88.7 &  41.04 \\
\hline
$\mathtt{LED}^\dag_{\mathtt{base}}$ & 56.2/26.6/28.8   & 24.2/4.5/15.4
 & 25.1/6.7/18.8 &	26.6 &	18.5 &	25.8/25.4 & 71.5 &  29.16 \\
$\mathtt{\modelname}$-$\mathtt{ED}$ & 49.8/22.2/25.8  & 30.8/7.1/18.6 & 28.3/6.7/18.7 & 33.1 & 14.2 & 29.4/26.4 & 81.4 & 31.41 \\ 
\bottomrule
\end{tabular}}
\caption{\textbf{Results on the SCROLLS benchmark compared to SotA.} We evaluate several models on the SCROLLS benchmark. $\dag$ indicates results reported on the public leaderboard.}
\label{tab:scrolls_full}
\end{table*}

\boldparagraph{Effect of document structure}
Our models are initially pre-trained on structured documents, and since most downstream tasks also maintain document structure, we aim to investigate the impact of such structures. To accomplish this, we selected two tasks, GovReport \citep{Huang2021NAACL} and QASPER \citep{Dasigi2021NAACL}, where document structure preservation is integral. We proceeded by flattening the documents to create a flattened version of the dataset. Subsequently, we trained both LED and \modelname-E on these datasets, employing the pseudo-section setting introduced in \secref{sec:language_tasks} for the flattened data. A comparative analysis between models trained with and without real document structure for the two datasets is presented in \tabref{tab:doc_structure}. While we observe that modeling document structure is not having a very large effect on downstream task performance in this setting, it is yet very important to utilize document structure during pre-training in our experiments. In fact, models pre-trained on pseudo sections and tested on the structured downstream tasks did not deliver any reasonable results. Our interpretation is that during pre-training, \modelname~learns to represent hierarchical information in different granularity via its anchor tokens, which allows it to adapt to pseudo-section data via fine-tuning.

\begin{table*}[t!]
\centering
\setlength{\tabcolsep}{5pt}
\begin{tabular}{l|c|ccccc}
\toprule
\multirow{3}{*}{\textbf{Model}}  & \textbf{GovRep} & \multicolumn{5}{|c}{\textbf{QASPER}}  \\
&   ROUGE-1/2/L     & \multicolumn{5}{|c}{F1} \\ 
& & Extractive & Abstractive & Yes/No & Unanswer. & Overall \\
\hline
$\mathtt{LED}$ & 56.2/26.6/28.8   & 30.96 &  15.76 & 70.33 & 26.21 &   32.80 \\
 \hline
$\mathtt{\modelname}$-$\mathtt{ED}$ & 49.89/21.54/25.26  & 30.57  &  11.42 & 67.14 & 46.45 & 33.14 \\ 
+ struct. & 49.42/21.22/24.97 & 33.12 & 13.02 & 64.19 & 43.17 & 34.02  \\
\bottomrule
\end{tabular}
\caption{\textbf{Effect of using document structure on GovRep and QASPER.} Note that the models are evaluated with the original GovRep~\citep{Huang2021NAACL} and QASPER~\citep{Dasigi2021NAACL} datasets, therefore results in this table might be slightly different from the results in \tabref{tab:scrolls} for the same task as the authors have cleaned the datasets in SCROLLS.}
\label{tab:doc_structure}
\end{table*}

\end{document}